\documentclass[10pt, twocolumn, twoside] {IEEEtran}		

\usepackage{graphicx}
\usepackage{amsmath}
\usepackage{amssymb}
\usepackage{amsthm}
\usepackage{booktabs}
\usepackage{cite}
\usepackage{url}
\usepackage{mathtools}
\usepackage{array}

\theoremstyle{definition}
\newtheorem{prop}{Proposition}
\newtheorem{corollary}{Corollary}
\newtheorem{defn}{Definition}

\newcommand{\tr}{^{T}}					
\newcommand{\reg}{\phi}					

\newcommand{\iter}[1]{^{(#1)}}

\newcommand{\opt}{^{\ast}}

\newcommand{\CS}{\mathcal{S}}		
\newcommand{\KK}{\mathcal{K}}		

\newcommand{\suchthat}{\mathrel{}\middle|\mathrel{}}  

\newcommand{\est}[1]{\hat{#1}} 

\DeclareMathOperator{\tf}{\theta}		

\newcommand{\act}[2]{ {#1}\iter{#2} }		

\newcommand{\std}{\mathrm{std}}

\DeclareMathOperator{\sign}{sign}

\providecommand{\norm}[1]{\lVert#1\rVert}
\providecommand{\abs}[1]{\left\lvert#1\right\rvert}

\newcommand{\inv}{^{-1}}

\newcommand{\lam}{{\lambda} }
\newcommand{\half}{\frac{1}{2}}

\newcommand{\eps}{\epsilon}

\newcommand{\h}{\mathbf{h}}

\renewcommand{\r}{\mathbf{r}}

\renewcommand{\u}{\mathbf{u}}
\renewcommand{\v}{\mathbf{v}}
\newcommand{\w}{\mathbf{w}}
\newcommand{\x}{\mathbf{x}}
\newcommand{\y}{\mathbf{y}}

\newcommand{\A}{\mathbf{A}}
\newcommand{\B}{\mathbf{B}}

\renewcommand{\H}{\mathbf{H}}
\newcommand{\I}{\mathbf{I}}

\newcommand{\R}{\mathbf{R}}

\newcommand{\0}{\mathbf{0}}

\newcommand{\calN}{\mathcal{N}}

\newcommand{\RR}{\mathbb{R}}

\newcommand{\ZZ}{\mathbb{Z}}

\newcommand{\ia}{({\it i\/})}
\newcommand{\ib}{({\it ii\/})}

\title{Sparse Signal Estimation by Maximally Sparse Convex Optimization }

\author{Ivan W. Selesnick and Ilker Bayram
\thanks{Copyright (c) 2013 IEEE. Personal use of this material is permitted. However, permission to use this material for any other purposes must be obtained from the IEEE by sending a request to pubs-permissions@ieee.org.}
\thanks{I. W. Selesnick is with the Department of Electrical and Computer Engineering, 
NYU Polytechnic School of Engineering,
6 Metrotech Center, Brooklyn, NY 11201, USA.
Email: selesi@poly.edu.
I. Bayram is with the Department of Electronics and Communication Engineering,
Istanbul Technical University, Maslak, 34469, Istanbul, Turkey. Email: ilker.bayram@itu.edu.tr. }%
\thanks{This research was support by the NSF under Grant No. CCF-1018020.}
}

\begin{document}
\maketitle

\begin{abstract}

This paper addresses the problem of sparsity penalized least squares
for applications in sparse signal processing, e.g. sparse deconvolution.
This paper aims to induce sparsity more strongly than L1 norm regularization,
while avoiding non-convex optimization.
For this purpose, 
this paper describes the design and use of non-convex penalty functions (regularizers)
constrained so as to ensure the convexity of the total cost function, F, to be minimized.
The method is based on parametric penalty functions,
the parameters of which are constrained to ensure convexity of F.
It is shown that optimal parameters can be obtained by semidefinite programming (SDP).
This maximally sparse convex (MSC) approach yields 
maximally non-convex sparsity-inducing penalty functions constrained
such that the total cost function, F, is convex.
It is demonstrated that iterative MSC (IMSC)
can yield solutions substantially more sparse than 
the standard convex sparsity-inducing approach, i.e., L1 norm minimization.

\end{abstract}

\section{Introduction}

In sparse signal processing, the $ \ell_1 $ norm has special significance \cite{Sparsity_ProcIEEE_2010, Bach_2012_now}. 
It is the convex proxy for sparsity.
Given the relative ease with which convex problems can be reliably solved,
the $ \ell_1 $ norm is a basic tool in sparse signal processing.
However, penalty functions that promote sparsity more strongly than the
$ \ell_1 $ norm yield more accurate results in many sparse signal
estimation/reconstruction problems.
Hence, numerous algorithms have been devised to solve non-convex
formulations of the sparse signal estimation problem.
In the non-convex case, generally only a local optimal solution can be ensured;
hence solutions are sensitive to algorithmic details. 

This paper aims to develop an approach that promotes sparsity
more strongly than the $ \ell_1 $ norm, but which attempts to avoid
non-convex optimization as far as possible.
In particular, 
the paper addresses ill-posed linear inverse problems of the form
\begin{equation}
	\label{eq:Fx}
	\arg \min_{\x \in \RR^N} \; \Bigl\{ F(\x) = \norm{ \y - \H \x}_2^2 + \sum_{n=0}^{N-1} \lam_n \reg_n(x_n) \Bigr\}
\end{equation}
where
$ \lam_n > 0 $ and 
$ \reg_n : \RR \to \RR $
are sparsity-inducing regularizers (penalty functions)
for $  n \in \ZZ_N = \{0, \dots, N-1 \} $.
Problems of this form arise in denoising, deconvolution, compressed sensing, etc.
Specific motivating applications include
nano-particle detection for bio-sensing
and
near infrared spectroscopic time series imaging
\cite{Selesnick_2012_TSP, Selesnick_2013_LPFTVD}.

This paper explores the use of non-convex penalty functions $ \reg_n $,
under the constraint that the total cost function $ F $ is convex
and therefore reliably minimized.
This idea, introduced by Blake and Zimmerman \cite{Blake_1987}, is carried out
\begin{quote}
\dots by balancing the positive second derivatives in the first term [quadratic fidelity term] 
against the negative second derivatives in the [penalty] terms \cite[page 132]{Blake_1987}.
\end{quote}
This idea is also proposed by Nikolova in Ref.~\cite{Nikolova_1998_ICIP} where it is used in the denoising of binary images.
In this work, to carry out this idea, 
we employ penalty functions parameterized by variables $a_n$, 
i.e., $ \reg_n(x) = \reg(x; a_n) $,
wherein the parameters $ a_n $ are selected so as to ensure convexity of the total cost function $ F $.
We note that in \cite{Blake_1987},
the proposed family of penalty functions are quadratic around the origin
and that all $ a_n $ are equal.
On the other hand, the penalty functions we utilize in this work are non-differentiable at the origin
as in \cite{Nikolova_2008_SIAM, Nikolova_2010_TIP}
(so as to promote sparsity)
and the $ a_n $ are not constrained to be equal.

A key idea is that the parameters $ a_n $ can be optimized
to make the penalty functions $ \reg_n $  maximally non-convex
(i.e., maximally sparsity-inducing),
subject to the constraint that $ F $ is convex.
We refer to this as the `maximally-sparse convex' (MSC) approach.
In this paper, 
the allowed interval for the parameters $ a_n $, 
to ensure $ F $ is convex, is obtained by formulating a semidefinite program (SDP) \cite{Antoniou_2007},
which is itself a convex optimization problem.
Hence, in the proposed MSC approach, the cost function $ F$ to be minimized
depends itself on the solution to a convex problem.
This paper also describes an iterative MSC (IMSC) approach that
boosts the applicability and effectiveness of the MSC approach.
In particular, IMSC extends MSC to the case where $ \H $ is rank deficient
or ill conditioned; e.g., overcomplete dictionaries and deconvolution of 
near singular systems.

The proposed MSC approach requires a suitable parametric penalty function $ \reg(\cdot\, ; a) $,
where $ a $ controls the degree to which $ \reg $ is non-convex.
Therefore, this paper also addresses the choice of parameterized non-convex penalty functions
so as to enable the approach.
The paper proposes suitable penalty functions $ \reg $ and describes their relevant properties.

\subsection{Related Work (Threshold Functions)}

When $ \H $ in \eqref{eq:Fx} is the identity operator, 
the problem is one of denoising and is separable in $ x_n $.
In this case, a sparse solution $ \x $ is usually
obtained by some type of threshold function, $ \tf : \RR \to \RR$.
The most widely used threshold functions are
the soft and hard threshold functions \cite{Donoho_1994_Biom}.
Each has its disadvantages, and many other thresholding functions
that provide a compromise of the soft and hard thresholding functions
have been proposed -- for example:
the firm threshold \cite{Gao_1997},
the non-negative (nn) garrote \cite{Gao_1998, Figueiredo_2001_TIP},
the SCAD threshold function \cite{Fan_2001_JASA, Zou_2008_AS},
and
the proximity operator of the $\ell_p$ quasi-norm ($0<p<1$) \cite{Lorenz_2007}.
Several penalty functions are unified by the two-parameter
formulas given in \cite{Atto_2011_SIViP, Gholami_2011_TSP},
wherein threshold functions are derived as proximity operators \cite{Combettes_2011_chap}.
(Table 1.2 of \cite{Combettes_2011_chap} lists the proximity operators of numerous functions.)
Further threshold functions are defined directly by their functional
form \cite{Zhang_1998_SPL, Zhang_2001_TNN, Zhao_2005_cnf}.

Sparsity-based nonlinear estimation algorithms can also be developed 
by formulating suitable non-Gaussian probability models that reflect
sparse behavior, and by applying Bayesian estimation techniques 
 \cite{Clyde_1999, Hyvarinen_1999, Nadarajah_2007_SP, Fadili_2005_TIP, Achim_2003_TGRS, Portilla_2003_TIP, Ji_2008_TSP}.
We note that, the approach we take below is essentially a deterministic one;
we do not explore its formulation from a Bayesian perspective.

This paper develops a specific threshold function 
designed so as to have the three properties advocated in \cite{Fan_2001_JASA}:
unbiasedness (of large coefficients), sparsity, and continuity. 
Further, the threshold function $\tf$ and its corresponding penalty function $\reg$
are parameterized by two parameters: the threshold $T$ and
the right-sided derivative of $\tf$ at the threshold, i.e. $\tf'(T^+)$,
a measure of the threshold function's sensitivity.
Like other threshold functions,
the proposed threshold function biases large $\abs{x_n}$ less than does the soft threshold function,
but is continuous unlike the hard threshold function.
As will be shown below, 
the proposed function is most similar to the threshold function (proximity operator)
corresponding to the logarithmic penalty, 
but it is designed to have less bias.
It is also particularly convenient in algorithms for solving \eqref{eq:Fx} that
do not call on the threshold function directly, but instead
call on the derivative of penalty function, $ \reg'(x)$,
due to its simple functional form.
Such algorithms include iterative reweighted least squares (IRLS) \cite{Harikumar_1996_cnf},
iterative reweighted $\ell_1$ \cite{Wipf_2010_TSP, Candes_2008_JFAP},
FOCUSS \cite{Rao_2003_TSP}, and 
algorithms derived using majorization-minimization (MM) \cite{FBDN_2007_TIP}
wherein the penalty function is upper bounded (e.g. by a quadratic or linear function).

\subsection{Related Work (Sparsity Penalized Least Squares)}
\label{sec:relatedB}

Numerous problem formulations and algorithms to obtain sparse solutions to the general ill-posed linear inverse problem, \eqref{eq:Fx},
have been proposed.
The $\ell_1$ norm penalty (i.e., $\reg_n(x) = \abs{x}$)
has been proposed for sparse deconvolution \cite{Kaaresen_1997_TSP, Claerbout_1973_Geo, Taylor_1979_Geo, Brien_1994_TSP}
and more generally for sparse signal processing \cite{Chen_1998_SIAM} and statistics \cite{Tibshirani_1996}.
For the $\ell_1$ norm and other non-differentiable convex penalties,
efficient algorithms for large scale problems of the form \eqref{eq:Fx}
and similar (including convex constraints)
have been developed based on proximal splitting methods \cite{Combettes_2008_SIAM, Combettes_2011_chap},
alternating direction method of multipliers (ADMM) \cite{Boyd_2011_admm},
majorization-minimization (MM) \cite{FBDN_2007_TIP},
primal-dual gradient descent \cite{Esser_2010},
and Bregman iterations \cite{Goldstein_2009_SIAM}.

Several approaches aim to obtain solutions to \eqref{eq:Fx} that are more sparse than the $\ell_1$ norm solution.
Some of these methods proceed first by selecting a non-convex penalty function that induces sparsity
more strongly than the $\ell_1$ norm,
and second by developing non-convex optimization algorithms for the minimization of $ F$;
for example, iterative reweighted least squares (IRLS)  \cite{Harikumar_1996_cnf, Wipf_2010_TSP},
FOCUSS \cite{Gorodnitsky_1997_TSP, Rao_2003_TSP},
extensions thereof \cite{Tan_2011_TSP, Mourad_2010_TSP},
half-quadratic minimization \cite{Charbonnier_1997_TIP, Geman_1995_TIP},
graduated non-convexity (GNC) \cite{Blake_1987},
and its extensions \cite{Nikolova_1998_TIP, Nikolova_1999_TIP, Nikolova_2008_SIAM, Nikolova_2010_TIP}.

The GNC approach for minimizing a non-convex function $ F $ proceeds by minimizing a sequence of approximate functions,
starting with a convex approximation of $ F $ and ending with $ F $ itself.
While GNC was originally formulated for image segmentation with smooth penalties,
it has been extended to general ill-posed linear inverse problems \cite{Nikolova_1998_TIP} 
and non-smooth penalties \cite{Nikolova_2008_SIAM, Nikolova_2010_TIP, Mohimani_2009_TSP}.

With the availability of fast reliable algorithms for $\ell_1$ norm minimization,
reweighted $\ell_1$ norm minimization is a suitable approach for the non-convex problem \cite{Candes_2008_JFAP, Wipf_2010_TSP}:
the tighter upper bound of the non-convex penalty provided by the weighted $\ell_1$ norm,
as compared to the weighted $\ell_2$ norm,
reduces the chance of convergence to poor local minima.
Other algorithmic approaches include `difference of convex' (DC) programming \cite{Gasso_2009_TSP}
and operator splitting \cite{Chartrand_2009_ISBI}.

In contrast to these works, 
in this paper the penalties $\reg_n$ are constrained by the operator $ \H $ and by $ \lambda_n $.
This approach (MSC) deviates from the usual approach wherein the penalty
is chosen based on prior knowledge of $\x$.
We also note that, by design,
the proposed approach leads to a convex optimization problem;
hence, it differs  from approaches that pursue non-convex optimization.
It also differs from usual convex approaches for sparse signal estimation/recovery
which utilize convex penalties.
In this paper, the aim is precisely to utilize non-convex penalties 
that induce sparsity more strongly than a convex penalty possibly can.

The proposed MSC approach is most similar to the generalizations of GNC to non-smooth
penalties \cite{Nikolova_1999_TIP, Nikolova_2008_SIAM, Nikolova_2010_TIP} that have proven effective
for the fast image reconstruction with accurate edge reproduction.
In GNC, the convex approximation of $ F $ is based on the minimum eigenvalue of $\H\tr \H$.
The MSC approach is similar but more general: not all $a_n$ are equal.
This more general formulation leads to an SDP, not an eigenvalue problem.
In addition, GNC comprises a sequence of non-convex optimizations, whereas the
proposed approach (IMSC) leads to a sequence of convex problems.
The GNC approach can be seen as a continuation method, wherein
a convex approximation of $ F $ is gradually transformed to $ F $ in a predetermined manner.
In contrast, in the proposed approach, each optimization problem is defined 
by the output of an SDP which depends on the support of the previous solution.
In a sense, $ F $ is redefined at each iteration, to obtain progressively sparse solutions.

By not constraining all $ a_n $ to be equal, the MSC approach allows a more general parametric form for the
penalty, and as such, it can be more non-convex (i.e., more sparsity promoting)
than if all $ a_n $ are constrained to be equal.
The example in Sec.~\ref{sec:deconv} compares the two cases (with and without the simplification that all $a_n$ are equal)
and shows that the simplified version gives inferior results. 
(The simplified form is denoted IMSC/S in Table \ref{table:deconv} and Fig. \ref{fig:deconv3} below).

If the measurement matrix $ \H $ is rank deficient, and if all $ a_n $ were constrained to be equal,
then the only solution in the proposed approach would have $ a_n = 0 $ for all $n$; i.e., the penalty function would be convex. 
In this case, it is not possible to gain anything by allowing the penalty function
to be non-convex subject to the constraint that the total cost function is convex.
On the other hand, the proposed MSC approach, depending on $ \H $,
can still have all or some $ a_n > 0 $ and hence can admit non-convex penalties
(in turn, promoting sparsity more strongly). 

\smallskip
\noindent
\textbf{L0 minimizaton:}
A distinct approach to obtain sparse solutions to \eqref{eq:Fx} is 
to find an approximate solution minimizing the $ \ell_0 $ quasi-norm
or satisfying an $\ell_0$ constraint.
Examples of such algorithms include:
matching pursuit (MP) and orthogonal MP (OMP) \cite{Mallat_1998},
greedy $\ell_1$ \cite{Kozlov_2010_geomath},
iterative hard thresholding (IHT) \cite{Blumensath_2010_STSP, Blumensath_2012_SP, Kingsbury_2003_ICIP,Portilla_2007_SPIE},
hard thresholding pursuit \cite{Foucart_2010_SIAM}, 
smoothed $\ell_0$, \cite{Mohimani_2009_TSP},
iterative support detection (ISD) \cite{Wang_2010_SIAM},
single best replacement (SBR) \cite{Soussen_2011_TSP},
and ECME thresholding \cite{Qiu_2012_TSP_ECME}.

Compared to algorithms aiming to solve the $ \ell_0 $ quasi-norm problem, 
the proposed approach again differs.
First, the $\ell_0$ problem is highly non-convex,
while the proposed approach defines a convex problem.
Second, methods for $\ell_0$ seek the correct
support (index set of non-zero elements) of $ \x $
and 
do not regularize (penalize) 
any element  $x_n$ in the calculated support.
In contrast, the design of the regularizer (penalty) is 
at the center of the proposed approach,
and no $x_n$ is left unregularized.

\section{Scalar Threshold Functions}

The proposed threshold function and corresponding penalty function is intended to
serve as a compromise between soft and hard threshold functions,
and as a parameterized family of functions for use with
the proposed MSC method for ill-posed linear inverse problems,
to be described in Sect.~\ref{sec:SPLS}.

First, we note the high sensitivity of the hard threshold function to small changes in its input.
If the input is slightly less than the threshold $T$, then a small positive perturbation produces a large change in the output,
i.e.,  $\tf_{\textup{h}}(T-\eps) = 0$ and  $\tf_{\textup{h}}(T+\eps) \approx T$
where $\tf_{\textup{h}} : \RR \to \RR $ denotes the hard threshold function.
Due to this discontinuity, spurious noise peaks/bursts often appear as a result of hard-thresholding denoising.
For this reason, a continuous threshold function is often preferred.
The susceptibility of a threshold function $\tf$ to the phenomenon of spurious noise peaks can
be roughly quantified by the maximum value its derivative attains, i.e., $\max_{y\in \RR} \tf'(y)$,
provided $\tf$ is continuous.
For the threshold functions considered below, $\tf'$ attains its maximum value at $ y = \pm T^{+} $;
hence, the value of $ \tf'(T^+) $ will be noted.
The soft threshold function $ \tf_{\textup{s}}$ has $ \tf_{\textup{s}}'(T^+) = 1 $
reflecting its insensitivity.
However, $\tf_{\textup{s}}$ substantially biases (attenuates) large values of its input; i.e.,
$ \tf_{\textup{s}}(y) = y - T $ for $ y > T $.

\subsection{Problem Statement}

In this section, we seek a threshold function and corresponding penalty 
\ia\
for which
the `sensitivity' $\tf'(T^+)$ can be readily tuned from 1 to infinity
and 
\ib\ 
that does not substantially bias large $y$,
i.e., $ y - \tf(y) $ decays to zero rapidly as $ y $ increases. 

For a given penalty function $ \reg $, the proximity operator \cite{Combettes_2011_chap}
denoted  $ \tf : \RR \to \RR $ is defined by
\begin{equation}
	\label{eq:deftf}
	\tf(y) =
	\arg \min_{x \in \RR} 	\;
	\left\{ F(x) = \half (y - x)^2 + \lam \reg(x) \right\}
\end{equation}
where $ \lam > 0 $.
For uniqueness of the minimizer, we assume in the definition of $ \tf(y) $ that $ F $ is strictly convex.
Common sparsity-inducing penalties include
\begin{equation}
	\label{eq:pfabslog}
	\reg(x) =  \abs{ x }
	\qquad
	\text{and}
	\qquad
	\reg(x) = \frac{1}{a} \log(1 + a \abs{x}).
\end{equation}
We similarly assume in the following that $ \reg(x) $ is three times continuously differentiable for all $ x \in \RR $ except $ x = 0 $,
and that $ \reg $ is symmetric, i.e., $ \reg(-x) = \reg(x)$.

If $ \tf(y)  = 0 $ for all $ \abs{y} \leqslant T $ for some $ T > 0 $,
and $ T $ is the maximum such value, 
then the function $\tf $ is a threshold function
and $ T $ is the threshold.

It is often beneficial in practice if $\tf$ admits a simple functional form.
However, 
as noted above, a number of algorithms for solving \eqref{eq:Fx}
do not use $\tf$ directly,  but use $\reg'$ instead.
In that case, 
it is beneficial if $\reg'$ has a simple function form.
This is relevant in Sec.~\ref{sec:SPLS} where such algorithms will be used.

In order that $ y - \tf(y) $ approaches zero, 
the penalty function $ \reg $ must be non-convex,
as shown by the following.

\begin{prop}
Suppose $\reg:\mathbb{R}\rightarrow \mathbb{R}$ is a convex function
and $\tf(y)$ denotes the proximity operator associated with $\reg$, defined in \eqref{eq:deftf}.
If $0 \leqslant  y_1 \leqslant  y_2$, then 
\begin{equation}
	\label{eqn:ineqprop}
	y_1 - \tf(y_1) \leqslant  y_2 - \tf(y_2).
\end{equation}
\begin{proof}
Let $u_i = \tf(y_i)$ for $i=1,2$. We have,
\begin{equation}
	\label{eq:yuu}
	y_i \in u_i + \lam \partial \reg(u_i).
\end{equation}
Since $y_2 \geqslant  y_1$, by the monotonicity of both of the terms on the right hand side of \eqref{eq:yuu},
it follows that $u_2 \geqslant  u_1$. 

If $u_2 = u_1$, \eqref{eqn:ineqprop} holds with since $y_2 \geq y_1$.

Suppose now that $u_2 > u_1$. Note that the subdifferential  $\partial \reg$ is also a monotone mapping since $\reg$ is a convex function. 
Therefore it follows that if $z_i \in \lam \partial \reg(u_i)$, we should have $z_2 \geqslant  z_1$.
Since $y_i - \tf(y_i) \in \lam \partial \reg(u_i)$, the claim follows.
\end{proof}
\end{prop}

According to the proposition, if the penalty is convex, then the gap between
$\tf(y)$ and $y$ increases as the magnitude of $y$ increases.
The larger $ y $ is, the greater the bias (attenuation) is.
The soft threshold function is an extreme case that keeps this gap constant (beyond the threshold $T$, the gap is equal to $T$).
Hence, in order to avoid attenuation of large values, the penalty function must be non-convex.

\subsection{Properties}

As detailed in the Appendix, 
the proximity operator (threshold function) $ \tf $ defined in \eqref{eq:deftf} can be expressed as
\begin{equation}
	\label{eq:tffinv}
	\tf(y) =
	\begin{cases}
		0, & \abs{y} \le T
		\\
		f\inv(y), \ & \abs{y} \ge T
	\end{cases}
\end{equation}
where the threshold, $ T $, is given by
\begin{equation}
	\label{eq:Treg}
	T = \lam \, \reg'(0^+)
\end{equation}
and $ f : \RR_+ \to \RR $ is defined as
\begin{equation}
	f(x) = x + \lam \reg'(x).
\end{equation}
As noted in the Appendix, 
 $ F$ is strictly convex if 
\begin{equation}
	\label{eq:regd2cc}
	\reg''(x) > -\frac{1}{\lam}, \quad \forall x > 0.
\end{equation}
In addition, we have
\begin{equation}
	\label{eq:dtf1}
	\tf'(T^+) = \frac{1}{1 + \lam \reg''(0^+)}
\end{equation}
and
\begin{equation}
	\label{eq:dtf2}
	\tf''(T^+) = - \frac{\lam \, \reg'''(0^+)}{[1+ \lam \reg''(0^+)]^3}.
\end{equation}
Equations \eqref{eq:dtf1} and \eqref{eq:dtf2} 
will be used in the following.
As noted above, $ \tf'(T^+) $ reflects the maximum sensitivity of $ \tf $.
The value $ \tf''(T^+) $ is also relevant; it will be set in Sec.~\ref{sec:atan}
so as to induce $ \tf(y)-y $ to decay rapidly to zero.

\subsection{The Logarithmic Penalty Function}

\begin{figure}[t]
	\centering
	\includegraphics{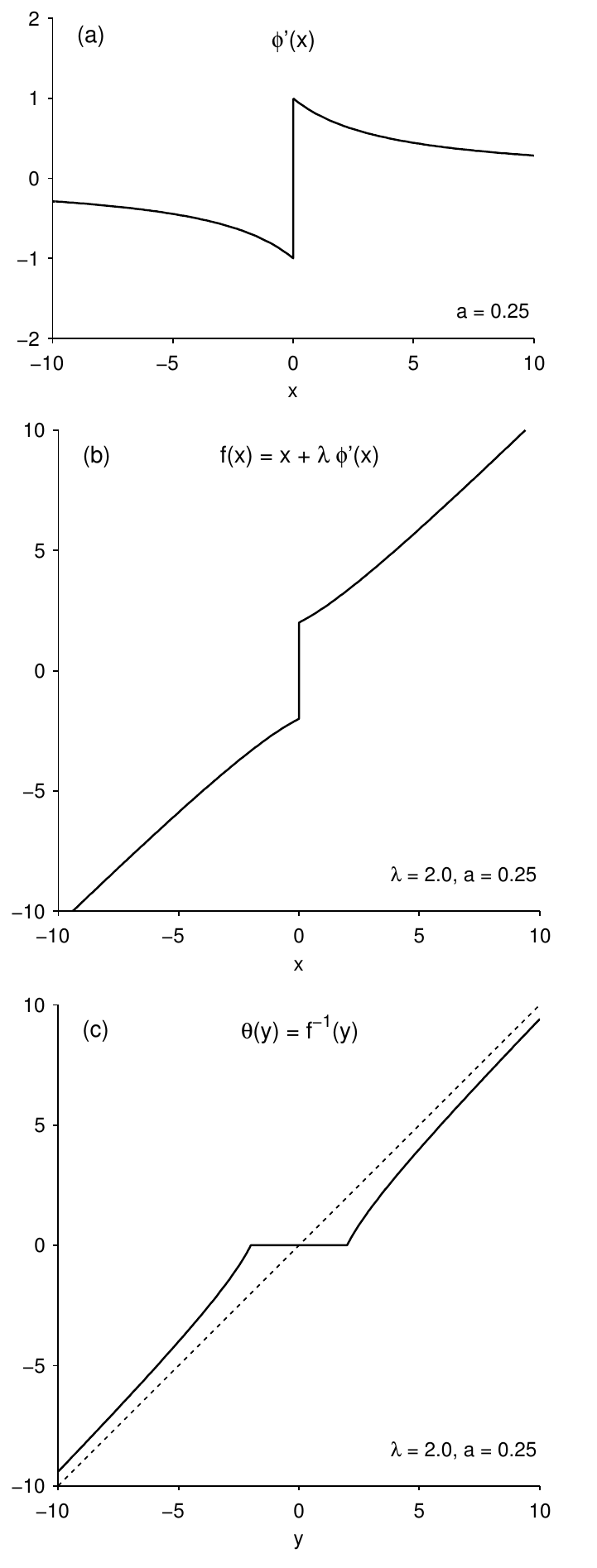}
	\caption{
		Functions related to the logarithmic penalty function \eqref{eq:log_reg}.
		(a) $\reg'(x)$.
		(b) $ f(x) = x + \lam \reg'(x) $.
		(c) Threshold function, $ \tf(y) = f\inv(y) $.
	}
	\label{fig:log}
\end{figure}

The logarithmic penalty function can be used for the MSC method
to be described in Sec.~\ref{sec:SPLS}.
It also serves as the model for the penalty function
developed in Sec.~\ref{sec:atan} below,
designed to have less bias.
The logarithmic penalty is given by
\begin{equation}
	\label{eq:log_reg}
		\reg(x) = \frac{1}{a} \log(1 + a \abs{x}),
		\quad
		0 < a \leq \frac{1}{\lam}
\end{equation}
which is differentiable except at $ x = 0 $.
For $ x \neq 0 $, the derivative of $\reg$ is given by 
\begin{equation}
	\label{eq:logdiff}
	\reg'(x) =
	\frac{1}{1 + a \abs{x}} \, \sign(x),  \quad x \neq 0,
\end{equation}
as illustrated in Fig.~\ref{fig:log}a.
The function $ f(x) = x + \lam \reg'(x) $ is
illustrated in Fig.~\ref{fig:log}b.
The threshold function $\tf$, given by \eqref{eq:tffinv}, is illustrated in Fig.~\ref{fig:log}c.

Let us find the range of $ a $ for which $ F $ is convex.
Note that
\[	\reg''(x) = - \frac{a}{(1 + a x)^2},
	\qquad
	\reg'''(x) = \frac{2 a^2}{(1 + a x)^3}
\]
for $x > 0$.
Using the condition \eqref{eq:regd2cc}, 
it is deduced that if $ 0 < a \leq 1/\lam $, then
$ f(x) $ is increasing, 
the cost function $ F $ in \eqref{eq:deftf} is convex,
and 
the threshold function $ \tf $ is continuous.

Using \eqref{eq:Treg},  the threshold is given by $T = \lam$.
To find $ \tf'(T^+) $ and $ \tf''(T^+) $, note that
$ \reg''(0^+) = -a $,
and
$ \reg'''(0^+) = 2 a^2$.
Using \eqref{eq:dtf1} and \eqref{eq:dtf2},
we then have
\begin{equation}
	\label{eq:log_dtf}
	\tf'(T^+) = \frac{1}{1 -  a \lam},
	\qquad
	\tf''(T^+) = - \frac{2 a^2 \lam}{(1 -  a \lam)^3}.
\end{equation}
As $ a $ varies between $ 0 $ and $ 1/\lam $,
the derivative $ \tf'(T^+) $ varies between $ 1 $ and infinity.
As $ a $ approaches zero, $ \tf $ approaches the soft-threshold function.
We can set $ a $ so as to specify $\tf'(T^+)$.
Solving \eqref{eq:log_dtf} for $ a $ gives
\begin{equation}
	\label{eq:aderiv}
	a = \frac{1}{\lam} \left(1 - \frac{1}{\tf'(T^+)} \right).
\end{equation}
Therefore, $ T $ and $ \tf'(T^+) $ can be directly 
specified by setting the parameters $ \lam $ and $ a $ (i.e., $ \lam = T $ and $ a $ is given by \eqref{eq:aderiv}).

Note that $ \tf''(T^+) $ given in \eqref{eq:log_dtf}
is strictly negative except when $ a = 0 $ which corresponds to the soft threshold function.
The negativity of $\tf''(T^+)$ inhibits the rapid approach of $ \tf $ to the identity function.

The threshold function $\tf$ is obtained by solving $y = f(x)$ for $ x $,
leading to
\begin{equation}
	a x^2 + (1-a \abs{y}) \abs{x} + (\lam - \abs{y}) = 0,
\end{equation}
which leads in turn to the explicit formula
\[
	\tf(y) =
	\begin{cases}
		\left[\frac{\abs{y}}{2}-\frac{1}{2a} + \sqrt{(\frac{\abs{y}}{2}+\frac{1}{2a})^2 - \frac{\lam}{a}}  \right] \sign(y),
						\quad	& \abs{y} \geqslant \lam \\
		0,				& \abs{y} \leqslant \lam \\
	\end{cases}
\]
as illustrated in Fig.~\ref{fig:log}c.
As shown, the gap $ y -  \tf(y) $ goes to zero for large $ y $.
By increasing $ a $ up to $1/\lam$, the gap goes to zero more rapidly;
however, increasing $ a $ also changes $ \tf'(T^+) $.
The single parameter $a$ affects both the derivative at the threshold and the convergence rate to identity.

The next section derives a penalty function, for which the gap goes to zero more
rapidly, for the same value of $ \tf'(T^+) $.
It will be achieved by setting $ \tf''(T^+) = 0 $.

\subsection{The Arctangent Penalty Function}
\label{sec:atan}

To obtain a penalty approaching the identity more rapidly
than the logarithmic penalty, we use equation \eqref{eq:logdiff}
as a model, and define a new penalty by means of its derivative as
\begin{equation}
	\reg'(x) = \frac{1}{b x^2 + a \abs{x} + 1} \, \sign(x),
	\qquad
	a > 0,
	\;
	b > 0.
\end{equation}
Using \eqref{eq:Treg}, the corresponding threshold function $\tf$
has threshold $ T = \lam. $ 
In order to use \eqref{eq:dtf1} and \eqref{eq:dtf2},
we note
\[
	\reg''(x) = -\frac{(2bx + a)}{(b x^2 + a x + 1)^2}
	\quad \text{for $x > 0$}
\]
\[
	\reg'''(x) = \frac{2 (2b x + a)^2}{(b x^2 + a x + 1)^3} - \frac{2 b}{ (b x^2 + a x + 1)^2 }
	\quad \text{for $x > 0$}.
\]
The derivatives at zero are given by
\begin{equation}
	\label{eq:regd}
	\reg'(0^+) = 1,
	\quad
	\reg''(0^+) = -a,
	\quad
	\reg'''(0^+) = 2 a^2 - 2  b.
\end{equation}
Using \eqref{eq:dtf1}, \eqref{eq:dtf2}, and \eqref{eq:regd}, we have
\begin{equation}
	\label{eq:tftp}
	\tf'(T^+) = \frac{1}{1 - \lam a},
	\quad
	\tf''(T^+) = \frac{2 \lam (b - a^2)}{(1 - \lam a)^3}.
\end{equation}
We may set $a$ so as to specify $ \tf'(T^+) $. 
Solving \eqref{eq:tftp} for $ a $ gives \eqref{eq:aderiv},
the same as for the logarithmic penalty function.

In order that the threshold function increases rapidly toward the identity function,
we use the parameter $ b $.
To this end, 
we set $ b $ so that $ \tf $ is approximately linear in the vicinity of the threshold.
Setting $ \tf''(T^+) = 0 $ in \eqref{eq:tftp} gives
$	b = a^2.  $
Therefore, the proposed penalty function is given by
\begin{equation}
	\reg'(x) = \frac{1}{a^2 x^2 + a \abs{x} + 1} \, \sign(x).
\end{equation}
From the condition \eqref{eq:regd2cc}, we find that if $ 0 < a \leq 1/\lam $, then $ f(x) = x + \lam \reg'(x) $ is strictly increasing, 
$ F $ is strictly convex, and $ \tf $ is continuous.
The parameters, $ a $ and $ \lam $, can be set as for the logarithmic penalty function;
namely $ T = \lam $ and by \eqref{eq:aderiv}.

To find the threshold function $\tf$, we
solve $ y = x + \lam \reg'(x) $ for $x$
which leads to
\begin{equation}
	a^2 \abs{x^3} + a (1-\abs{y}a) x^2 + (1-\abs{y}a) \abs{x} + (\lam - \abs{y}) = 0
\end{equation}
for $ \abs{y} > T $.
The value of $ \tf(y) $ can be found solving the cubic polynomial for $x$,
and multiplying the real root by $\sign(y)$.
Although $\tf$ does not have a simple functional form,
the function $ \reg'$ does.
Therefore, algorithms such as MM and IRLS, which use $ \reg' $ instead of $ \tf $,
can be readily used in conjunction with this penalty function.

The penalty function itself, $\reg$, can be found by integrating its derivative:
\begin{align}
	\reg(x) & = 
	\int_0^{\abs{x}} \reg'(u) \, du
	\\
	\label{eq:atan_reg}
	& = 
	 \frac{2}{a\sqrt{3}}  \, \left( 
	 	\tan\inv\left( \frac{1+2 a \abs{x}}{\sqrt{3}}\right)
		-
		\frac{\pi}{6}
	\right).
\end{align}
We refer to this as the arctangent penalty function.

\begin{figure}
	\centering
	\includegraphics{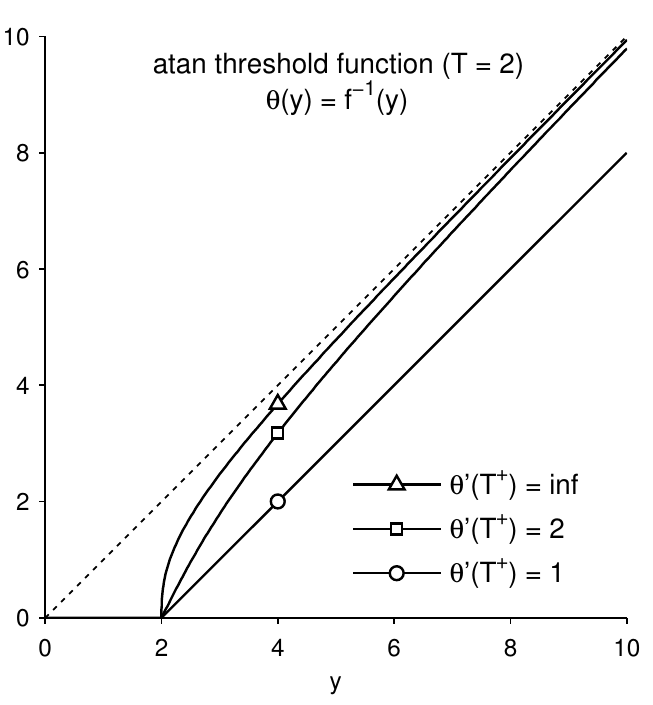}
	\caption{
		The arctangent threshold function for several values of $\tf'(T^+)$.
	}
	\label{fig:atan}
\end{figure}

The threshold function is illustrated in Fig.~\ref{fig:atan}
for threshold $ T = \lam = 2 $ and 
three values of $ \tf'(T^+) $.
With $ \lam = 2 $, the function $ F $ is strictly convex for all $ a \in [0, 1/\lam] $.
With $ \tf'(T^+) = 1 $, one gets $ a = 0 $ and 
$ \tf $ is the soft-threshold function.
With $ \tf'(T^+) = 2 $, one gets $ a = 1/4 $ and 
$ \tf $ converges to the identity function.
With $ \tf'(T^+) = \infty $, one gets $ a = 1/2 $;
in this case, $ \tf $ converges more rapidly to the identity function,
but $\tf$ may be more sensitive than desired 
in the vicinity of the threshold.

\begin{figure}[t]
	\centering
	\includegraphics{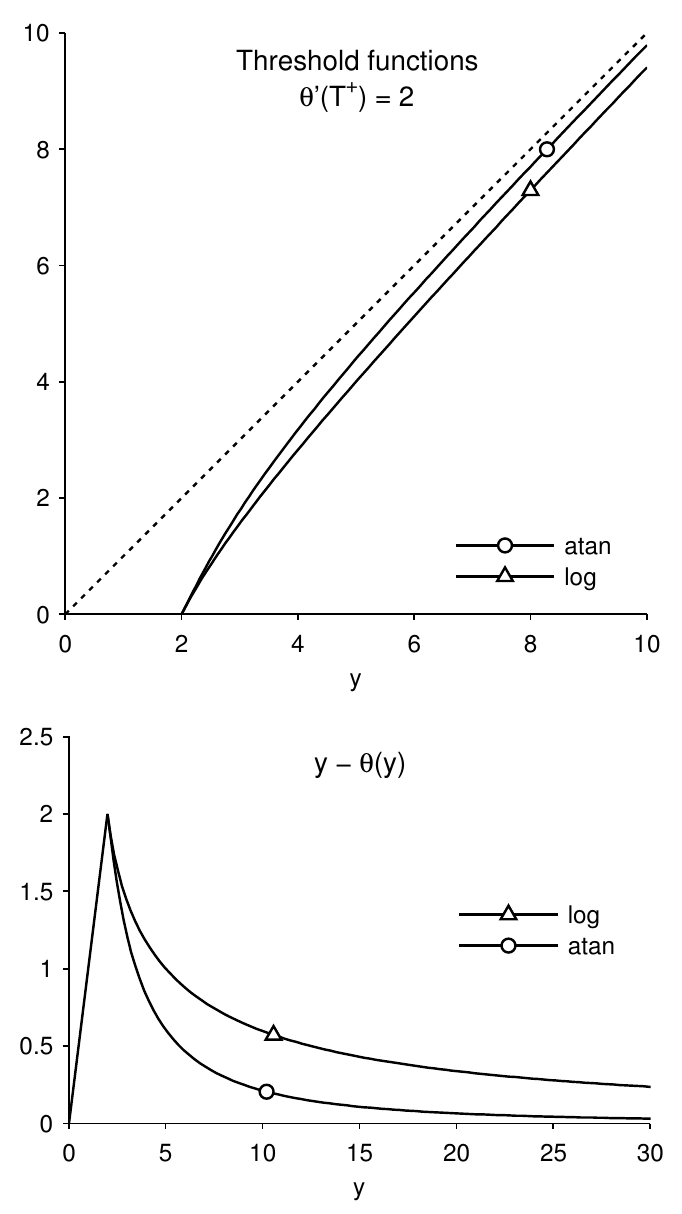}
	\caption{
		Comparison of arctangent and logarithmic penalty functions, both with $\tf'(T^+) = 2$.
		The arctangent threshold function 
		approaches the identity faster than the logarithmic penalty function.
	}
	\label{fig:atanlog}
\end{figure}

Figure~\ref{fig:atanlog} compares the logarithmic and arctangent
threshold functions where the parameters for each function
are set so that $T$ and $ \tf'(T^+) $ are the same,
specifically, $ T = \tf'(T^+) = 2 $.  
It can be seen that the arctangent threshold function converges
more rapidly to the identity function than the logarithmic threshold function.
To illustrate the difference more clearly,
the lower panel in Fig.~\ref{fig:atanlog} 
shows the gap between the identity function and the threshold function.
For the arctangent threshold function, this gap goes to zero more rapidly.
Yet, for both threshold functions, $\tf'$ has a maximum value of 2.
The faster convergence of the arctangent threshold function 
is due to $\reg'(x)$ going to zero like $ 1/x^2 $,
whereas for the logarithmic threshold function $\reg'(x)$ goes to zero like $ 1/x $.

\begin{figure}
	\centering
	\includegraphics{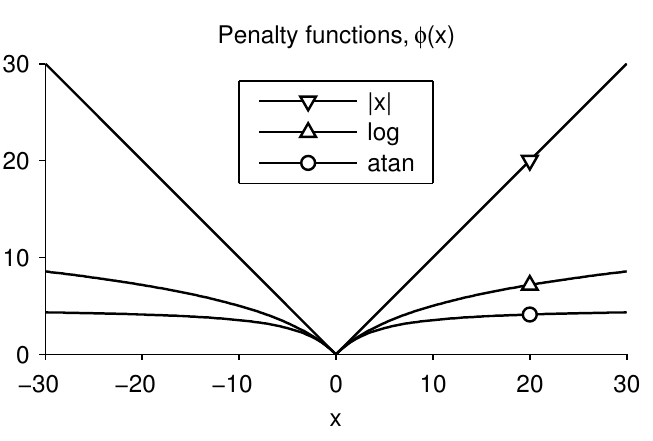}
	\caption{
	Sparsity promoting penalties: absolute value ($\ell_1$ norm), logarithmic, and arctangent 
	penalty functions ($ a = 0.25 $).
	}
	\label{fig:penalties}
\end{figure}

Figure~\ref{fig:penalties} compares the logarithmic and arctangent penalty functions.
Both functions grow more slowly than $ \abs{x} $ and thus induce less bias
than the $ \ell_1 $ norm for large $ x $.
Moreover, while the logarithmic penalty tends to $+\infty$, 
the arctangent penalty tends to a constant. 
Hence, the arctangent penalty leads to less bias than the logarithmic penalty.
All three penalties have the same slope (of 1) at $ x = 0 $;
and furthermore, the logarithmic and arctangent penalties have the same 
second derivative (of $-a$) at $ x = 0 $.
But, the logarithmic and arctangent penalties have different third-order derivatives at $ x = 0 $
($2a^2$ and zero, respectively).
That is, the arctangent penalty is more concave at the origin than the logarithmic penalty.

\subsection{Other Penalty Functions}

The firm threshold function \cite{Gao_1997},
and the smoothly clipped absolute deviation (SCAD) threshold function \cite{Fan_2001_JASA, Zou_2008_AS}
also provide a compromise between hard and soft thresholding.
Both the firm and SCAD threshold functions are continuous and equal to the identity function for large $\abs{y}$
(the corresponding $ \reg'(x) $ is equal to zero for $ x $ above some value).
Some algorithms, such as IRLS, MM, etc., involve dividing by $ \reg' $,
and for these algorithms, divide-by-zero issues arise.
Hence, the penalty functions corresponding to these threshold functions are unsuitable for these algorithms.
 
A widely used penalty function is the $\ell_p$ pseudo-norm, $ 0 < p < 1$,
for which $ \reg(x) = \abs{x}^p. $
However, using \eqref{eq:regd2cc},
it can be seen that for this penalty function,
the cost function $F(x)$ is not convex for any $ 0 < p < 1 $.
As our current interest is in non-convex penalty functions for which
$F$ is convex, we do not further discuss the $\ell_p$ penalty. 
The reader is referred to \cite{Lorenz_2007, Nikolova_2005_SIAM} for in-depth analysis of this 
and several other penalty functions.

\subsection{Denoising Example}

\begin{figure}
	\centering
	\includegraphics{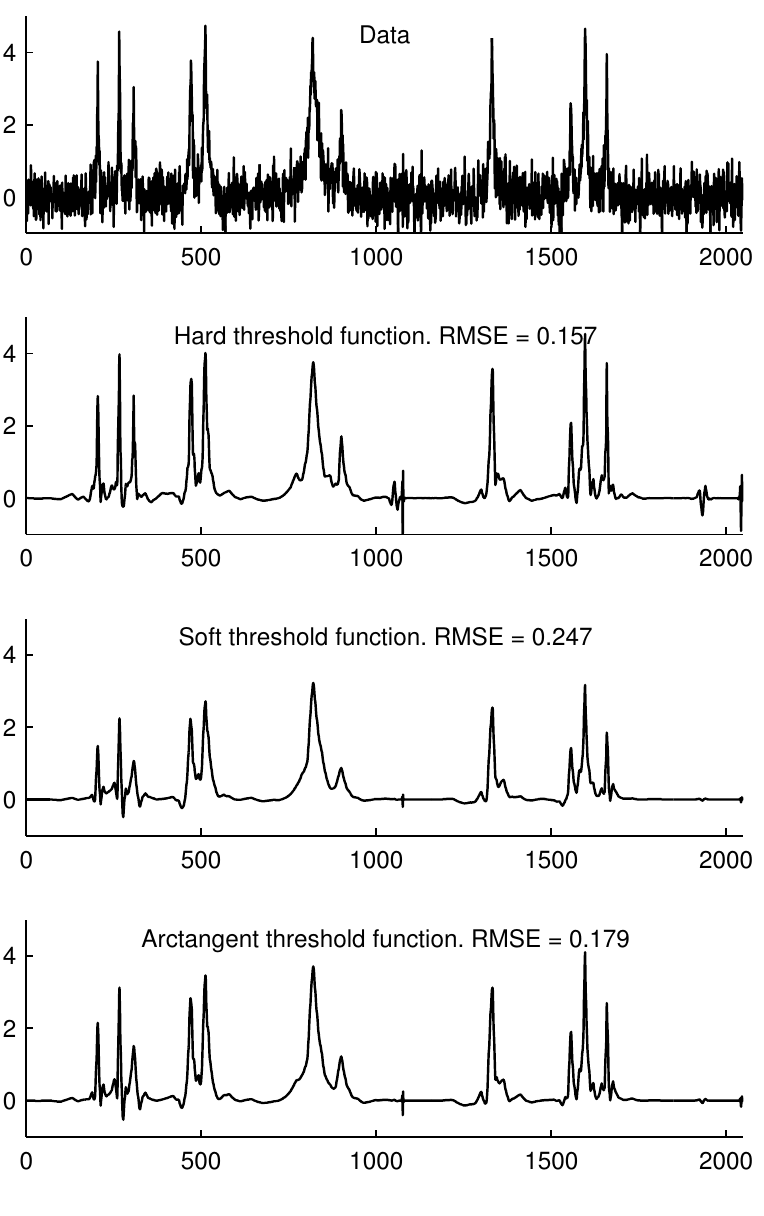}
	\caption{
		Denoising via orthonormal wavelet thresholding using
		various threshold functions. 
	}
	\label{fig:dnoise}
\end{figure}

To illustrate the trade-off between $ \tf'(T^+) $ and the bias introduced by thresholding,
we consider the denoising of the noisy signal illustrated in Fig.~\ref{fig:dnoise}.
Wavelet domain thresholding is performed with several thresholding functions.

Each threshold function is applied with the same threshold, $ T = 3 \sigma $.
Most of the noise (c.f. the `three-sigma rule') will fall below the
threshold and will be eliminated.
The RMSE-optimal choice of threshold is usually lower than $3 \sigma$,
so this represents a larger threshold than that usually used.
However, a larger threshold reduces the number of spurious noise peaks
produced by hard thresholding.

The hard threshold achieves the best RMSE, but
the output signal exhibits spurious noise bursts due to
noisy wavelet coefficients exceeding the threshold.
The soft threshold function reduces the spurious noise bursts,
but attenuates the peaks and results in a higher RMSE.
The arctangent threshold function suppresses the noise bursts,
with modest attenuation of peaks, and 
results in an RMSE closer to that of hard thresholding.

In this example, the signal is `bumps' from WaveLab \cite{wavelab}, with length 2048.
The noise is additive white Gaussian noise with standard deviation $\sigma = 0.4 $.
The wavelet is the orthonormal Daubechies wavelet with 3 vanishing moments.

\section{Sparsity penalized least squares}
\label{sec:SPLS}

Consider the linear model,
\begin{equation}
	\label{eq:model}
	\y = \H \x + \w
\end{equation}
where $\x \in \RR^N$ is a sparse $N$-point signal,
$ \y \in \RR^M $ is the observed signal,
$ \H \in \RR^{M \times N} $ is a linear operator (e.g., convolution),
and
$\w \in \RR^N $ is additive white Gaussian noise (AWGN).
The vector $\x$ is denoted $\x = (x_0, \dots, x_{N-1})\tr $.

Under the assumption that $ \x $ is sparse, we consider the 
linear inverse problem:
\begin{equation}
	\label{eq:defF}
	\arg \min_{\x \in \RR^N}
	\biggl\{
	F(\x) = 
	\half \norm{ \y - \H \x }_2^2 + \sum_{n=0}^{N-1} \lam_n \reg( x_n; a_n)
	\biggr\}
\end{equation}
where $ \reg( x; a ) $ is a sparsity-promoting penalty function with parameter $ a$,
such as the logarithmic or arctangent penalty functions.
In many applications, all $\lam_n $ are equal, i.e., $ \lam_n = \lam $.
For generality, we let this regularization parameter  depend on the index $n$.

In the following, we address the question of 
how to constrain the regularization parameters $ \lam_n $ and $ a_n $ 
so as to ensure $ F $ is convex, 
even when $ \reg( \, \cdot \,; a_n ) $ is not convex.
A problem of this form is addressed in GNC \cite{Blake_1987, Nikolova_1999_TIP},
where the $ a_n $ are constrained to be equal.

\subsection{Convexity Condition}

Let $ \reg(x; a)$ denote a penalty function with parameter $a$.
Consider the function $v : \RR \to \RR $, defined as
\begin{equation}
	\label{eq:deflilf}
	v(x) = \half x^2 + \lam \, \reg(x; a).
\end{equation}
Assume $v(x)$ can be made strictly convex
for special choices of $\lambda$ and $a$.
We give a name to the set of all such choices.
\begin{defn}
Let $ \CS $ be the set of pairs $(\lam, a)$ for which $v(x)$ in \eqref{eq:deflilf} is strictly convex.
We refer to $ \CS $ as the `parameter set associated with $\reg$'.
\end{defn}

For the logarithmic and arctangent penalty functions
described above, the set $ \CS $ is given by
\begin{equation}\label{eqn:special}
	\CS = \{ (\lam, a) :  \lam > 0, \; 0 \leqslant a \leqslant 1/\lam \}.
\end{equation}
Now, consider  the function $ F : \RR^N \to \RR $, defined in \eqref{eq:defF}.
The following proposition provides a sufficient condition on $ (\lam_n, a_n) $ ensuring the strict convexity of $ F $.
\begin{prop}  
Suppose $ \R $ is a positive definite diagonal matrix such that $ \H\tr \H - \R$ is positive semidefinite.
Let $r_n$ denote the $n$-th diagonal entry of $\R$,
i.e., 
$	[\R]_{n,n} = r_n > 0.  $
Also, let $ \CS $ be the parameter set associated with $\reg$.
If $ (\lam_n/r_n, a_n) \in \CS $ for each $n$,
then $F(\x)$ in \eqref{eq:defF} is strictly convex.
\begin{proof}
The function $ F(\x) $ can be written as
\begin{equation}
	F(\x) = 
	\underbrace{ \half \x\tr (\H\tr \H - \R) \x - \y\tr \H \x + \half \y\tr \y }_{q(\x)} + g(\x),
\end{equation}
where
\begin{equation}
	g(\x) = \half \x\tr \R \x +  \sum_n \lam_n \reg( x_n; a_n).
\end{equation}
Note that $q(\x)$ is convex since $\H\tr \H - \R$ is positive semidefinite.
Now, since $\R$ is diagonal, we can rewrite $g(\x)$ as
\begin{align}
	\label{eq:rnlamA}
	g(\x)	
	& =
	\sum_n \frac{r_n}{2} \, x_n^2 + \lam_n \reg( x_n; a_n) 
	\\
	\label{eq:rnlam}
	& =
	\sum_n r_n \Big( \half \, x_n^2 + \frac{\lam_n}{r_n} \reg( x_n; a_n)  \Bigr).
\end{align}
From \eqref{eq:rnlam}, it follows that if $ (\lam_n/r_n, a_n) \in \CS $ for each $n$, then $ g(\x) $ is strictly convex.
Under this condition, being a sum of a convex and a strictly convex function, it follows that $F(\x)$ is strictly convex.
\end{proof}
\end{prop}

The proposition states that constraints on the penalty parameters $ a_n $
ensuring strict convexity of $ F(\x) $ 
can be obtained using a diagonal matrix $\R$ lower bounding $ \H\tr \H$.
If $ \H $ does not have full rank, then strict convexity is precluded.
In that case, $\H\tr\H$ will be positive semidefinite.
Consequently, $\R$ will also be positive semidefinite, with some $r_n$ equal to zero.
For those indices $n$, where $r_n=0$, the quadratic term in \eqref{eq:rnlamA} vanishes.
In that case, we can still ensure the convexity of $F$ in \eqref{eq:defF}
by ensuring $\reg(x;a_n)$ is convex.
For the logarithmic and arctangent penalties proposed in this paper, we have $\reg(x; a) \to \abs{x}$ as $ a \to 0 $.
Therefore, we define $\reg(x; 0) = \abs{x} $ for the log and atan penalties.
 
In view of \eqref{eqn:special}, the following is a corollary of this result.

\begin{corollary}
For the logarithmic and arctangent penalty functions,
if 
\begin{equation}
	\label{eq:arlamc}
	0 < a_n < \frac{r_n}{\lam_n},
\end{equation}
then $ F $ in \eqref{eq:defF} is strictly convex.
\qed
\end{corollary}

\begin{figure}
	\centering
	\includegraphics{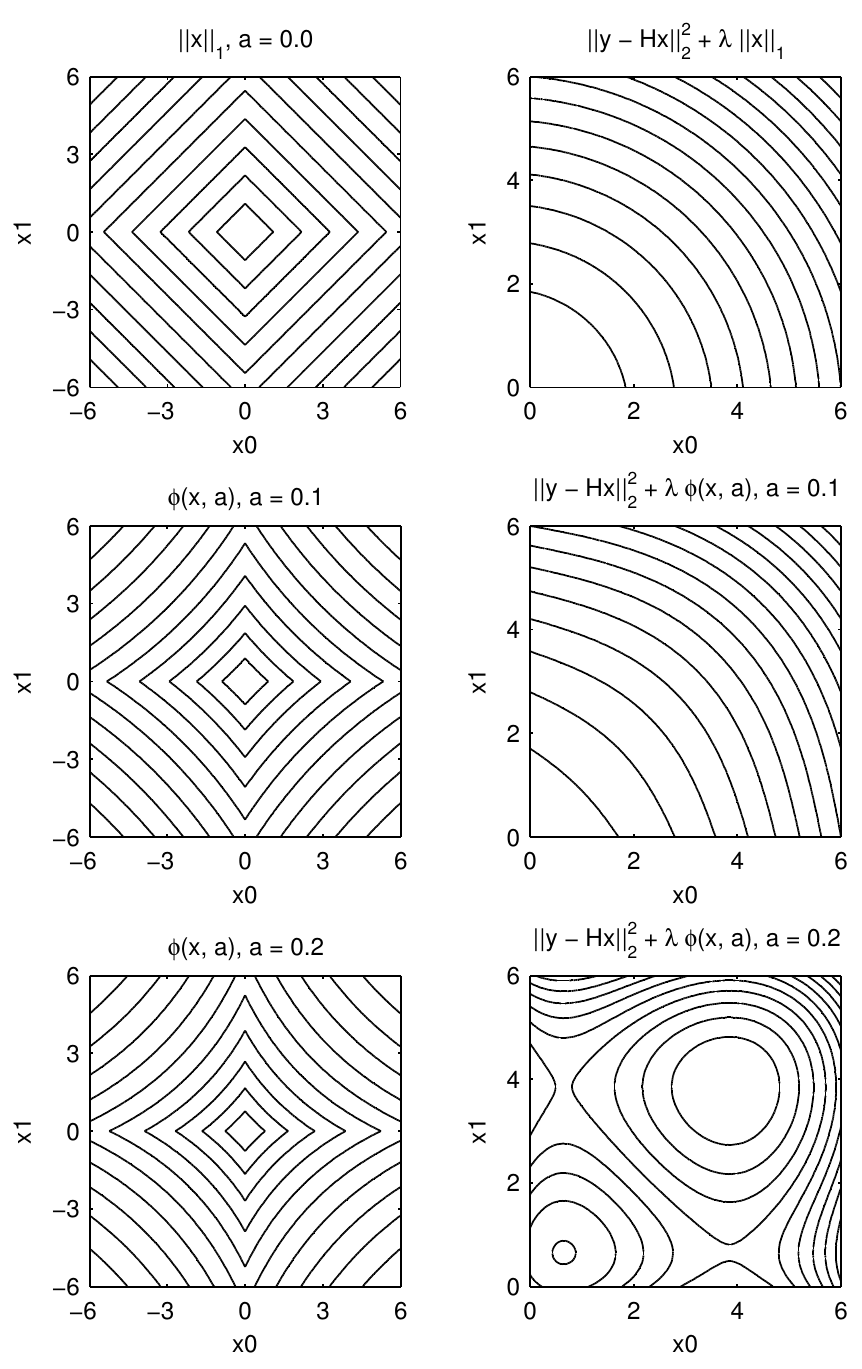}
	\caption{
		Contour plots of the logarithmic penalty function $\reg$ and cost function $ F $ 
		for three values of $ a $ as described in the text.
		For $ a = 0.1 $, the function $ F $ is convex even though the penalty function is not.
	}
	\label{fig:contours}
\end{figure}

We illustrate condition \eqref{eq:arlamc} with a simple example
using $ N = 2 $ variables.
We set $ \H = \I $,  $ \y = [9.5, 9.5]\tr $,  and $ \lam_0 = \lam_1 = 10$.
Then $ \R = \I  $ is a positive diagonal matrix with $ \H\tr \H - \R $ positive semidefinite.
According to \eqref{eq:arlamc},
$ F $ is strictly convex if $ a_i < 0.1 $, $ i = 0, 1 $.
Figure \ref{fig:contours} illustrates 
the contours of the logarithmic penalty function and the cost function $ F $ for
three values of $ a $.
For $ a = 0$, the penalty function reduces to the $\ell_1$ norm.
Both the penalty function and $ F $ are convex. 
For $ a = 0.1$, the penalty function is non-convex but $ F $ is convex.
The non-convexity of the penalty
is apparent in the figure (its contours do not enclose convex regions).
The non-convex `star' shaped contours induce sparsity more strongly than 
the diamond shaped contours of the $\ell_1$ norm.
For $ a = 0.2$, both the penalty function and $ F $ are non-convex.
The non-convexity of $ F $ is apparent in the figure (a convex function can not
have more than one stationary point, while the figure shows two).
In this case, the star shape is too pronounced for $ F $ to be convex.
In this example, $ a = 0.1 $ yields the maximally sparse convex (MSC) problem.

Can a suitable $ \R $ be obtained by variational principles?
Let us denote the minimal eigenvalue of $ \H\tr \H $ by  $ \alpha_\textup{min} $.
Then $ \R = \alpha_\textup{min} \I $
is a positive semidefinite diagonal lower bound, as needed.
However, this is a sub-optimal lower bound in general.
For example, if $ \H $ is a non-constant diagonal matrix, then 
a tighter 
lower bound is $ \H\tr \H $ itself, which
is very different from  $ \alpha_\textup{min} \I $. 
A tighter lower bound is of interest because the tighter the bound,
the more non-convex the penalty function can be,
while maintaining convexity of $ F $.
In turn, sparser solutions can be obtained without 
sacrificing convexity of the cost function.
A tighter lower bound can be found as the solution to an optimization problem,
as described in the following.

\subsection{Diagonal Lower Bound Matrix Computation}

Given $ \H $, the convexity conditions above calls for a
positive semidefinite diagonal matrix $\R$ lower bounding $\H\tr \H$.
In order to find a reasonably tight lower bound, each $ r_n $ should be maximized.
However, these $N$ parameters must be chosen jointly to ensure $ \H\tr \H - \R $ is positive semidefinite.
We formulate the calculation of  $ \R $ as an optimization problem:
\begin{equation}
	\label{eq:sd}
	\begin{split}
	\arg \max_{\r \in \RR^N}
	\quad & \sum_{n=0}^{N-1} r_n
	\\
	\text{such that} \quad &
		r_n \geqslant \alpha_{\textup{min}}
	\\
	& 
	\H\tr \H - \R \geqslant 0
	\end{split}
\end{equation}
where $ \R $ is the diagonal matrix $ [\R]_{n,n} = r_n $.
The 
inequality $ \H\tr \H - \R \geqslant 0 $
expresses the constraint that $ \H\tr \H - \R $ is positive semidefinite (all its eigenvalues non-negative).
Note that the problem is feasible, because $ \R = \alpha_{\textup{min}} \I $ satisfies the constraints.
We remark that
problem \eqref{eq:sd} is not the only approach to derive a matrix $ \R $ satisfying  Proposition 2.
For example, the objective function could be a weighted sum or other norm of $ \{r_n\} $.
One convenient aspect of $\eqref{eq:sd}$ is that it has the form of a standard convex problem.

Problem \eqref{eq:sd} can be recognized as a semidefinite optimization problem,
a type of convex optimization problem for which algorithms have been developed
and for which software is available \cite{Antoniou_2007}.
The cost function in \eqref{eq:sd} is a linear function of the $N$ variables,
and the constraints are linear matrix inequalities (LMIs).
To solve \eqref{eq:sd} and obtain $\R$, we have used the MATLAB software 
package `SeDuMi' \cite{sedumi}.

Often, inverse problems arising in signal processing involve large data sets
(e.g., speech, EEG, and images).
Practical algorithms must be efficient in terms of memory and computation.
In particular, they should be `matrix-free',
i.e., the operator $\H$ is not explicitly stored as a matrix,
nor are individual rows or columns of $\H$ accessed or modified.
However, optimization algorithms for semidefinite programming usually
involve row/column matrix operations and are not `matrix free'.
Hence,
solving problem \eqref{eq:sd} will likely be a bottleneck
for large scale problems.
(In the deconvolution example below, the MSC solution using SDP 
takes from 35 to 55 times longer to compute than the $\ell_1$ norm solution).
This motivates the development of semidefinite algorithms to solve \eqref{eq:sd}
where $ \H $ is not explicitly available, 
but for which multiplications by $ \H $ and $ \H\tr $ are fast
(this is not addressed in this paper).

Nevertheless, for 1D problems of `medium'-size (arising for example in biomedical applications \cite{Selesnick_2012_TSP}),
\eqref{eq:sd} is readily solved via existing software.
In case \eqref{eq:sd} is too computationally demanding, 
then the suboptimal choice  $ \R = \alpha_{\textup{min}} \I $ 
can be used as in GNC \cite{Blake_1987, Nikolova_1999_TIP}.
Furthermore, we describe below a multistage algorithm
whereby the proposed MSC approach is applied iteratively.

\subsection{Optimality Conditions and Threshold Selection}

When the cost function $F$ in \eqref{eq:defF} is strictly convex, then its minimizer
must satisfy specific conditions \cite{Fuchs_2004_Tinfo}, \cite[Prop 1.3]{Bach_2012_now}.
These conditions can be used to verify the optimality of
a solution produced by a numerical algorithm.
The conditions also aid in setting the regularization parameters $ \lam_n $.

If $F$ in \eqref{eq:defF} is strictly convex, 
and $\reg$ is differentiable except at zero,
then $\x\opt$ minimizes $ F $ if
\begin{equation}
	\label{eq:optim}
	\begin{cases}
		\displaystyle
		\frac{1}{\lam_n} [\H\tr (\y - \H \x\opt)]_n =  \reg'(x_n\opt; a_n), 	\quad & x_n\opt \neq 0	\\[1em]
		\displaystyle
		\frac{1}{\lam_n} [\H\tr (\y - \H \x\opt)]_n \in  [\reg'(0^-; a_n), \, \reg'(0^+; a_n)], 	\qquad & x_n\opt = 0
	\end{cases}
\end{equation}
where $ [\v]_n $ denotes the $n$-th component of the vector $ \v $.

The optimality of a numerically obtained solution can be illustrated
by a scatter plot of $   [\H\tr (\y - \H \x)]_n/\lam_n $ versus $  x_n a_n $, for $ n \in \ZZ_N $.
For the example below, Fig.~\ref{fig:deconv2} illustrates
the scatter plot, wherein the points lie on the graph of $ \reg' $.
We remark that the scatter plot representation as in Fig.~\ref{fig:deconv2} 
makes sense only when the parametric penalty $ \reg(x; a)$ is a function
of $ a x $ and $ a $,
as are the log and atan penalties, \eqref{eq:log_reg} and \eqref{eq:atan_reg}.
Otherwise, the horizontal axis will not be labelled $x_n a_n$ and
the points will not lie on the graph of $ \reg $.
This might not be the case for other parametric penalties.

The condition \eqref{eq:optim} can be used to set the regularization parameters, $ \lam_n $,
as in Ref.~\cite{Fuchs_2009_sysid}.
Suppose $\y$ follows the model \eqref{eq:model} where $ \x $ is sparse. 
One approach for setting $ \lam_n $ is to ask that the solution to \eqref{eq:defF} be all-zero when $ \x $ is all-zero
in the model \eqref{eq:model}.
Note that, if $ \x = \0 $, then $ \y $ consists of noise only (i.e.,\ $\y = \w$).
In this case, \eqref{eq:optim} suggests that $ \lam_n $ be chosen such that
\begin{equation}
	\label{eq:lamzopt}
	\lam_n \, \reg'(0^-; a_n) \leqslant [\H\tr \w]_n \leqslant \lam_n \, \reg'(0^+; a_n),
	\
	n \in \ZZ_N. 
\end{equation}
For the $ \ell_1 $ norm, logarithmic and arctangent penalty functions,
$ \reg(0^-; a_n)  = -1 $ and $ \reg(0^+; a_n)  = 1 $, so 
\eqref{eq:lamzopt} can be written as
\begin{equation}
	\label{eq:Atw}
	\abs{ [\H\tr \w]_n } \leqslant \lam_n,
	\quad
	n \in \ZZ_N. 	
\end{equation}
However, the larger $ \lam_n $ is, the more $ x_n $ will be attenuated.
Hence, it is reasonable to set $ \lam_n $ to the smallest value satisfying \eqref{eq:Atw},
namely,
\begin{equation}
	\label{eq:setlammax}
	\lam_n \approx \max \, \abs{ [ \H\tr \w]_n}
\end{equation}
where $ \w $ is the additive noise.
Although \eqref{eq:setlammax} assumes availability of the noise signal $\w$,
which is unknown in practice, \eqref{eq:setlammax} can often be
estimated based on knowledge of statistics of the noise $ \w $.
For example, based on the `three-sigma rule', we obtain
\begin{equation}
	\label{eq:setlamstd}
	\lam_n
	\approx 
	3 \, \std( [ \H\tr \w ]_n ).
\end{equation}
If $ \w $ is white Gaussian noise with variance $ \sigma^2 $, then 
\begin{equation}
	\std( [ \H\tr \w ]_n ) = 
	\sigma \norm{\H(\cdot, n)}_2
\end{equation}
where $ \H(\cdot, n) $ denotes column $n$ of $ \H $.
For example, 
if $ \H $ denotes linear convolution, then all columns of $ \H $ have
equal norm and \eqref{eq:setlamstd} becomes
\begin{equation}
	\label{eq:setlamconv}
	\lam_n = \lam \approx 3 \sigma \norm{\h}_2
\end{equation}
where $ \h $ is the impulse of the convolution system.

\subsection{Usage of Method}

We summarize the forgoing approach, MSC, to sparsity penalized least squares,
cf. \eqref{eq:defF}.
We assume the parameters $ \lam_n $ are fixed
(e.g., set according to additive noise variance).

\begin{enumerate}
\item
Input: $ \y \in \RR^M $, $ \H \in \RR^{M \times N} $, $ \{ \lam_n > 0, \; n \in \ZZ_N \} $, $ \reg : \RR\times \RR \to \RR $.

\item
Find a positive semidefinite diagonal matrix $ \R $ such that $ \H\tr \H - \R $ is positive semidefinite;
i.e., solve \eqref{eq:sd},
or use the sub-optimal $ \R = \alpha_{\textup{min}} \I $.
Denote the diagonal elements of $ \R $  by $ r_n, \; n \in \ZZ_N $.

\item
For $ n \in \ZZ_N $, set $ a_n $ such that $ (r_n / \lam_n , a_n) \in \CS $.
Here, $ \CS $ is the set such that $ v $ in \eqref{eq:deflilf} is convex if $ (\lam, a) \in \CS $.

\item
Minimize \eqref{eq:defF} to obtain $ \x $.

\item
Output: $ \x \in \RR^N $.
\hfill
$ \square $

\end{enumerate}

The penalty function $ \reg $ need not be the logarithmic or arctangent penalty functions 
discussed above. 
Another parametric penalty function can be used, but it must have the property that
$ v $ in \eqref{eq:deflilf} is convex for $ (\lam, a) \in \CS $ for some set $ \CS $.
Note that $ \reg(x, p) = \abs{x}^p $ with $ 0 < p < 1 $ does not qualify
because $ v $ is non-convex for all $ 0 < p < 1 $.
On the other hand, the firm penalty function \cite{Gao_1997} could be used.

In step (3), for the logarithmic and arctangent penalty functions, one can use
\begin{equation}
	a_n = \beta \frac{r_n}{\lam_n},
	\quad
	\text{where} \;
	0 \leqslant \beta \leqslant 1.
\end{equation}
When $ \beta = 0 $, the penalty function is simply the $\ell_1$ norm;
in this case, the proposed method offers no advantage
relative to $\ell_1$ norm penalized least squares (BPD/lasso).
When $ \beta = 1 $, the penalty function is maximally non-convex 
(maximally sparsity-inducing) subject to $ F $ being convex.
Hence, as it is not an arbitrary choice, $\beta = 1$ can be taken as a recommended default value.
We have used $ \beta = 1 $ in the examples below.

The minimization of \eqref{eq:defF} in step (4)
is a convex optimization problem for which numerous
algorithms have been developed as noted in Sec. \ref{sec:relatedB}.
The most efficient algorithm depends primarily on
the properties of $ \H $.

\subsection{Iterative MSC (IMSC)}

An apparent limitation of the proposed approach, MSC, is that for some problems of interest,
the parameters $ r_n $ are either equal to zero or nearly equal to zero for all $ n \in \ZZ_N $,
i.e., $ \R \approx \0 $.
In this case,  the method requires that $ \reg( \cdot\, ; a_n) $
be convex or practically convex.
For example, for the logarithmic and arctangent penalty functions,
$ r_n \approx 0 $ leads to $ a_n \approx 0 $.
As a consequence, the penalty function is practically the $ \ell_1 $ norm.
In this case, the method offers no advantage in comparison with
$ \ell_1 $ norm penalized least squares (BPD/lasso).

The situation wherein $ \R \approx \0 $ arises in two standard sparse signal processing problems:
basis pursuit denoising and deconvolution.
In deconvolution, if the system is non-invertible or nearly singular (i.e., the frequency response has 
a null or approximate null at one or more frequencies), 
then the lower bound $ \R $ will be $ \R \approx \0 $.
In BPD, the matrix $ \H $ often represents the inverse of an overcomplete frame (or dictionary),
in which case the lower bound $ \R $ is again close to zero.

In order to broaden the applicability of MSC, 
we describe iterative MSC (IMSC) wherein MSC is applied several times.
On each iteration, MSC is applied only to the non-zero elements of 
the sparse solution $\x$ obtained as a result of the previous iteration.
Each iteration involves only those columns of $ \H $ corresponding to
the previously identified non-zero components.
As the number of active columns of $ \H $ diminishes as the iterations progress,
the problem \eqref{eq:sd} produces a sequence of increasingly
positive diagonal matrices $\R$.
Hence, as the iterations progress, the penalty functions become increasingly non-convex.
The procedure can be repeated until there is no change in the index set of non-zero elements.

The IMSC algorithm can be initialized with the $\ell_1$ norm solution, i.e., using $ \reg(x, a_n) = \abs{x} $
for all $ n \in \ZZ_N $.
(For the logarithmic and arctangent penalties, $ a_n = 0, n \in \ZZ_N $.)
We assume the $ \ell_1 $ norm solution is reasonably sparse;
otherwise, sparsity is likely not useful for the problem at hand.
The algorithm should be terminated when there is no change (or only insignificant change) between
the active set from one iteration to the next.

The IMSC procedure is described as follows, where $i \geqslant 1$ denotes the iteration index.

\begin{enumerate}
\item
Initialization.
Find the $\ell_1$ norm solution:
\begin{equation}
	\label{eq:L1sol}
	\x\iter{1} = \arg \min_{\x \in \RR^N} \norm{ \y - \H \x }_2^2
		+ \sum_{n = 0}^{N-1} \lam_n \abs{ x_n}.
\end{equation}
Set $ i = 1 $ and $ K\iter{0} = N $.
Note $ \H $ is of size $ M \times N $.
\smallskip

\item
Identify the non-zero elements of $ \x\iter{i} $,
and record their indices in the set $ \KK\iter{i} $,
\begin{equation}
	\KK\iter{i} = \left\{ n \in \ZZ_N \suchthat x_n\iter{i} \neq 0 \right\}.
\end{equation}
This is the support of $ \x\iter{i} $.
Let $ K\iter{i} $ be the number of non-zero elements of $ \x\iter{i} $, i.e., $ K\iter{i} = \abs{\KK\iter{i}} $.
\smallskip

\item
Check the termination condition:
If $ K\iter{i} $ is not less than $ K\iter{i-1} $, then terminate.
The output is $ \x\iter{i} $.
\smallskip

\item
Define $\act{\H}{i} $ as the sub-matrix of $ \H $ containing only columns $ k \in \KK\iter{i} $.
The matrix $ \act{\H}{i} $ is of size $ M \times K\iter{i} $.

Find a positive semidefinite diagonal matrix $\R\iter{i}$ lower bounding $ [\act{\H}{i}]\tr \act{\H}{i} $,
i.e., 
solve problem \eqref{eq:sd}
or
use $\alpha_{\textup{min}}\iter{i} \I$.
The matrix $\R\iter{i}$ is of size $K\iter{i} \times K\iter{i}$.

\smallskip

\item
Set $ a_n $ such that $ (\lam_n/r_n\iter{i}, a_n ) \in \CS, \; n \in \KK\iter{i} $.
For example, with the logarithmic and arctangent penalties,
one may set
\begin{equation}
	a_n\iter{i} = \beta \frac{r_n\iter{i}}{\lam_n}, \quad n \in \KK\iter{i}
\end{equation}
for some $ 0 \leqslant \beta \leqslant 1 $.
\item
Solve the $K\iter{i}$ dimensional convex problem:
\begin{equation}
	\label{eq:optu}
	\u\iter{i} = \arg \min_{\u \in \RR^{K\iter{i}}} \norm{ \y - \act{\H}{i} \u }_2^2
		+ \sum_{n \in \KK\iter{i}} \lam_n \reg( u_n; a_n\iter{i} ).
\end{equation}

\item
Set $ \x\iter{i+1} $ as
\begin{equation}
	x\iter{i+1}_n = 
	\begin{cases}
		0,	&		n \notin \KK\iter{i}
		\\
		u\iter{i}_n,		&	n \in \KK\iter{i}.
	\end{cases}
\end{equation}

\item
Set $ i = i + 1 $
and
go to step 2).
\hfill
$ \square $

\end{enumerate}
\smallskip

In the IMSC algorithm, the support of $ \x\iter{i} $ 
can only shrink from one iteration to the next, i.e.,  $ \KK\iter{i+1} \subseteq \KK\iter{i} $
and $ K\iter{i+1} \leqslant K\iter{i}$. 
Once there is no further change in $\KK\iter{i}$, each subsequent iteration will produce exactly the same result,
i.e.,  
\begin{equation}
	\KK\iter{i+1} = \KK\iter{i}  \implies \x\iter{i+1} = \x\iter{i}.
\end{equation}
For this reason, the procedure should be terminated when $ \KK\iter{i} $ ceases to shrink.
In the 1D sparse deconvolution example below,
the IMSC procedure terminates after only three or four iterations.

Note that the problem \eqref{eq:sd} in step 4) reduces in size as the algorithm progresses.
Hence each instance of \eqref{eq:sd} requires less computation than the previous.
More importantly, 
each matrix $ \H\iter{i+1} $ has a subset of the columns of $ \H\iter{i} $.
Hence, $\R\iter{i+1}$ is less constrained than $\R\iter{i}$,
and 
the penalty functions become more non-convex (more strongly sparsity-inducing)
as the iterations progress.
Therefore, the IMSC algorithm produces a sequence of successively sparser $ \x\iter{i} $.

Initializing the IMSC procedure with the $\ell_1$ norm solution substantially
reduces the computational cost of the algorithm.
Note that if the $\ell_1$ norm solution is sparse, i.e., $ K\iter{1} \ll N $,
then all the semidefinite optimization problems \eqref{eq:sd}
have far fewer variables than $N$,
i.e., $ K\iter{i} \leqslant K\iter{1} $.
Hence, IMSC can be applied to larger data sets than would otherwise be computationally practical,
due to the computational cost of \eqref{eq:sd}.

\subsection{Deconvolution Example}
\label{sec:deconv}

\begin{figure}
	\centering
	\includegraphics{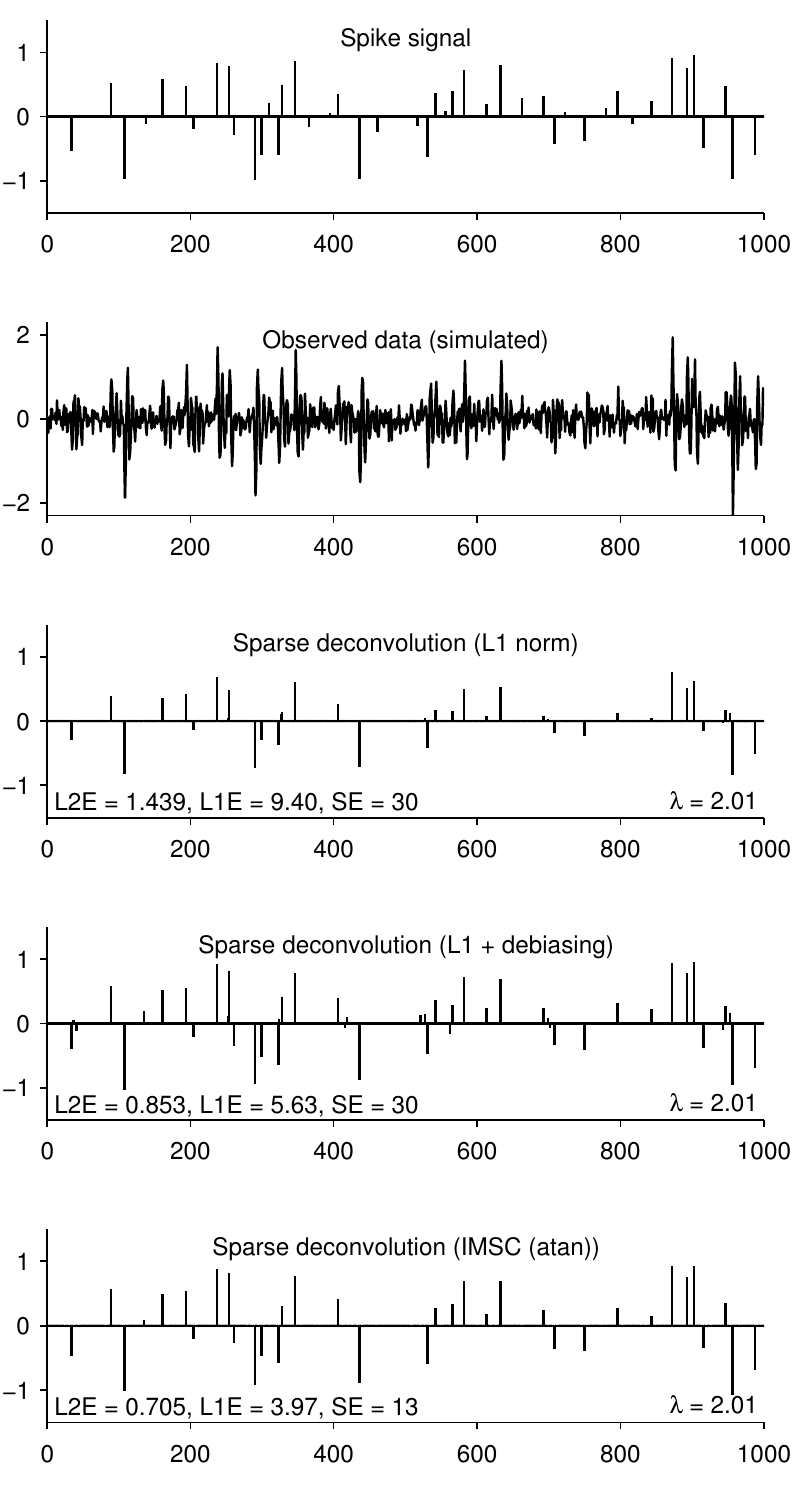}
	\caption{
		Sparse deconvolution via sparsity penalized least squares.
	}
	\label{fig:deconv}
\end{figure}

A sparse signal $x(n)$ of length $N = 1000$ is generated so that
\ia\ the inter-spike interval is uniform random between 5 and 35 samples,
and
\ib\ the amplitude of each spike is uniform between $-1$ and $1$.
The signal is illustrated in Fig.~\ref{fig:deconv}.

The spike signal is then used as the input to a linear time-invariant (LTI) system,
the output of which is contaminated by AWGN, $ w(n) $.
The observed data, $y(n)$, is written as
\[
	y(n) = \sum_k b(k) \, x(n-k) - \sum_k a(k) \, y(n-k) + w(n)
\]
where $ w(n) \sim \calN(0, \sigma^2) $.
It can also be written as
\[
	\y = \A\inv \B \x + \w = \H \x + \w,
	\quad
	\H = \A\inv \B
\]
where $ \A $ and $ \B $ are banded Toeplitz matrices \cite{Selesnick_2012_cnx_deconv}.
In this example, we set
$
	b(0) = 1, \; b(1) = 0.8, \; a(0) = 1, \; a(1) = -1.047, \; a(2) = 0.81,
$
and $ \sigma = 0.2 $.
The observed data, $\y$, is illustrated in Fig.~\ref{fig:deconv}.

Several algorithms for estimating the sparse signal $\x$ will be compared.
The estimated signal is denoted $\est{\x}$.
The accuracy of the estimation is 
quantified by the $\ell_2$ and $\ell_1$ norms of the error signal and
by the support error,
denoted L2E, L1E, and SE respectively.
\begin{enumerate}
\item
L2E $=\norm{\x - \est{\x}}_2 $
\item
L1E $=\norm{\x - \est{\x}}_1 $
\item
SE $ = \norm{ s(\x) - s(\est{\x})}_0 $
\end{enumerate}
The support error, SE, is computed using $s(\x)$, the $\eps$-support of $ \x \in \RR^N $.
Namely,
 $ s : \RR^N \to \{0, 1\}^N $ is defined as
\begin{equation}
	[s(\x)]_n =
	\begin{cases}
		1, \quad & \abs{x_n} > \eps
		\\
		0,	& \abs{x_n} \leqslant \eps
	\end{cases}
\end{equation}
where $ \eps > 0 $ is a small value to accommodate negligible non-zeros.
We set $ \eps = 10^{-3} $.
The support error, SE, counts both the false zeros and the false non-zeros of $ \est{\x} $.
The numbers of false zeros and false non-zeros are denoted FZ and FN, respectively.

First, the sparse $ \ell_ 1 $ norm solutions, i.e., $ \reg(x, a) = \abs{x} $ in \eqref{eq:defF},
with and without debiasing, are computed.\footnote{Debiasing is a post-processing step wherein least squares is performed over the obtained support set \cite{Figueiredo_2007_GPSR}.}
We set $ \lam_n $ according to \eqref{eq:setlamconv}, i.e., $ \lam_n = 2.01, \; n \in \ZZ_N$.
The estimated signals are illustrated in Fig.~\ref{fig:deconv}.
The errors L2E, L1E, and SE, are noted in the figure.
As expected, 
debiasing substantially improves the L2E and L1E errors of the $ \ell_1$ norm solution;
however, it does not improve the support error, SE.
Debiasing does not make the solution more sparse.
The errors, averaged over 200 trials, are shown in Table \ref{table:deconv}.
Each trial consists of independently generated sparse and noise signals.

\begin{table}
	\caption{
		Sparse deconvolution example. 
		Average errors (200 trials).
	}
	\label{table:deconv}
\centering
\begin{tabular}{@{}lccc@{\hspace{0.5em}  (}r@{$\ $}r@{}}
\toprule
Algorithm						&	L2E             &	L1E             &	SE &FZ,&FN)	\\
\midrule
$\ell_1$ norm                       		&	1.443           &	10.01           &	37.60      &10.3,&27.3)              \\ 
$\ell_1$ norm     + debiasing     	&	0.989           &	7.14            &	37.57      &10.3,&27.2)              \\ 
AIHT \cite{Blumensath_2012_SP}    &	1.073           &	6.37            &	24.90      &12.4,&12.5)              \\ 
ISD \cite{Wang_2010_SIAM}            &	0.911           &	5.19            &	19.67      &11.6,&8.1)              \\ 
SBR  \cite{Soussen_2011_TSP}       &	0.788           &	4.05            &	13.62      &12.0,&1.6)              \\ 
$\ell_p (p=0.7)$ IRL2                    	&	0.993           &	5.80            &	16.32      &12.9,&3.4)              \\ 
$\ell_p (p=0.7)$ IRL2    + debiasing &	0.924           &	4.82            &	16.32      &12.9,&3.4)              \\ 
$\ell_p (p=0.7)$ IRL1                    	&	0.884           &	5.29            &	14.43      &11.5,&2.9)              \\ 
$\ell_p (p=0.7)$ IRL1    + debiasing &	0.774           &	4.18            &	14.43      &11.5,&2.9)              \\ 
IMSC (log)                    			&	0.864           &	5.08            &	17.98      &9.8, & 8.2)          \\ 
IMSC (log)    + debiasing     		&	0.817           &	4.83            &	17.98      &9.8, & 8.2)          \\ 
IMSC (atan)                   			&	0.768           &	4.29            &	15.43      &10.0, & 5.5)          \\ 
IMSC (atan)   + debiasing     		&	0.769           &	4.35            &	15.42      &10.0, & 5.5)          \\ 
IMSC/S (atan)                 			&	0.910           &	5.45            &	17.93      &9.8, & 8.1)          \\ 
IMSC/S (atan) + debiasing     		&	0.800           &	4.73            &	17.92      &9.8, & 8.1)          \\ 
\bottomrule
\end{tabular}
\end{table}

Next, sparse deconvolution is performed using three algorithms developed
to solve the highly non-convex $\ell_0$ quasi-norm problem, 
namely the Iterative Support Detection (ISD) algorithm \cite{Wang_2010_SIAM},%
\footnote{{http://www.caam.rice.edu/\%7Eoptimization/L1/ISD/}}
the Accelerated Iterative Hard Thresholding (AIHT) algorithm \cite{Blumensath_2012_SP},%
\footnote{{http://users.fmrib.ox.ac.uk/\%7Etblumens/sparsify/sparsify.html}}
and the Single Best Replacement (SBR) algorithm \cite{Soussen_2011_TSP}.
In each case, we used software by the respective authors.
The ISD and SBR algorithms require regularization parameters $\rho$ and $\lam$ respectively;
we found that $\rho = 1.0 $ and $ \lam = 0.5 $ were approximately optimal.
The AIHT algorithm requires the number of non-zeros be specified;
we used the number of non-zeros in the true sparse signal.
Each of ISD, AIHT, and SBR significantly improve the accuracy
of the result in comparison with the $\ell_1$ norm solutions,
with SBR being the most accurate.
These algorithms essentially seek the correct support.
They do not penalize the values in the detected support;
so, debiasing does not alter the signals produced by these algorithms.

The $\ell_p$ quasi-norm, with $ p = 0.7 $, i.e. $\reg(x) = \abs{x}^p$, also substantially 
improves upon the $\ell_1$ norm result.
Several methods exist to minimize the cost function $ F $ in this case.
We implement two methods: IRL2 and IRL1 (iterative reweighted $\ell_2$ and $\ell_1$ norm minimization, respectively),
with and without debiasing in each case.
We used $ \lam = 1.0 $, which we found to be about optimal on average for this deconvolution problem.
As revealed in Table \ref{table:deconv}, IRL1 is more accurate than IRL2.
Note that IRL2 and IRL1 seek to minimize exactly the same cost function;
so the inferiority of IRL2 compared to IRL1 is due to the convergence of IRL2
to a local minimizer of $ F $.
Also note that debiasing substantially improves L2E and L1E (with no effect on SE) for both IRL2 and IRL1.
The $\ell_p$ results demonstrate both the value of a non-convex regularizer and the vulnerability of 
non-convex optimization to local minimizers.

The results of the proposed iterative MSC (IMSC) algorithm, with and without debiasing,
are shown in Table \ref{table:deconv}.
We used  $ \beta = 1.0 $ and $ \lam_n = 2.01, \; n \in \ZZ_N$,
in accordance with \eqref{eq:setlamconv}.
Results using the logarithmic (log) and arctangent (atan) penalty functions are tabulated,
which show the improvement provided by the later penalty, in terms of L2E, L1E, and SE.
While debiasing reduces the error (bias) of the logarithmic penalty,
it has negligible effect on the arctangent penalty.
The simplified form of the MSC algorithm, wherein
$ \R = \alpha_{\textup{min}} \I $ is used instead of the $ \R $ computed
via SDP, is also tabulated in Table \ref{table:deconv}, denoted by IMSC/S.
IMSC/S is more computationally efficient than MSC due to the omission of SDP;
however, it does lead to an increase in the error measures.

The IMSC algorithm ran for three iterations on average.
For example, the IMSC solution illustrated in Fig.~\ref{fig:deconv}
ran with $ K\iter{1} = 61$,  $ K\iter{2} = 40 $, and $ K\iter{3} = 38$.
Therefore, even though the signal is of length 1000,
the SDPs that had to be solved are much smaller: of sizes 61, 40, and 38, only.

\begin{figure}
	\centering
	\includegraphics{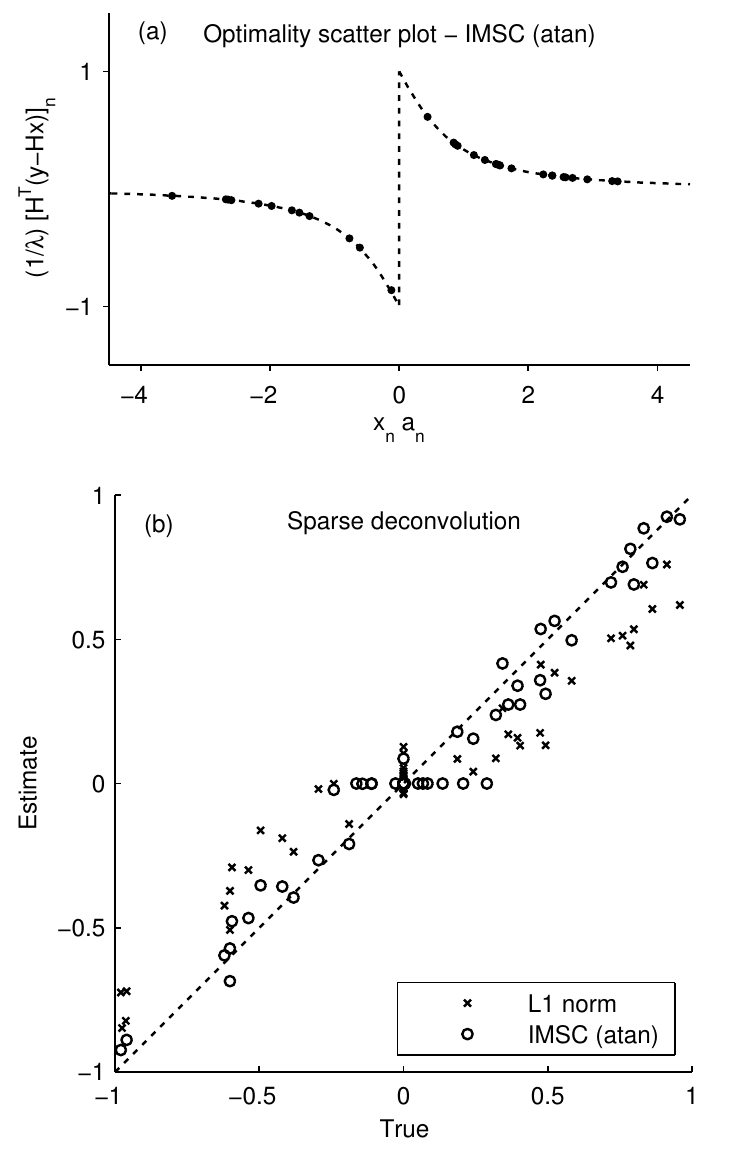}
	\caption{
		Sparse deconvolution.
		(a) Illustration of optimality condition \eqref{eq:optim} for IMSC (atan) solution.
		(b) Comparison of $\ell_1$ norm and IMSC solutions.
	}
	\label{fig:deconv2}
\end{figure}

The optimality of the MSC solution at each stage can be verified using \eqref{eq:optim}.
Specifically, a scatter plot of $   [\H\tr (\y - \H \x)]_n/\lam_n   $ verses $ x_n a_n $,
for all $ n \in \KK\iter{i} $,
should show all points lying on the graph of $ \partial \reg(x, 1) $.
For the IMSC solution illustrated in Fig.~\ref{fig:deconv},
this optimality scatter plot is illustrated in Fig.~\ref{fig:deconv2},
which shows that all points lie on the graph of $ \sign(x)/(1 + \abs{x} + x^2) $,
hence verifying the optimality of the obtained solution.

To more clearly compare the relative bias of the $\ell_1$ norm and IMSC (atan) solutions,
these two solutions are illustrated together in Fig.~\ref{fig:deconv2}.
Only the non-zero elements of each solution are shown.
In this figure, the closer the points lie to the identity, 
the more accurate the solution.
The figure shows the IMSC solution lies closer to the 
identity than the $\ell_1$ norm solution;
and 
the $\ell_1$ norm solution tends to underestimate the true values.

In terms of L2E and L1E, the best IMSC result, i.e., IMSC (atan), is outperformed by SBR and the IRL1 + debiasing algorithm.
In addition,  IMSC (atan) yields lower SE than $\ell_1$ minimization, AIHT, and ISD.
IMSC does not yield the best error measures, but it comes reasonably close;
even though IMSC is based entirely on convex optimization.
In terms of L1E and SE, the SBR performs best for this example.
Most notably, SBR attains a small number of false non-zeros.

Note that IMSC requires only the parameter $\beta$ (with $ 0 \leqslant \beta \leqslant 1 $) beyond those
parameters (namely $\lam_n$) required for the $\ell_1 $ norm solution.

\begin{figure}
	\centering
	\includegraphics{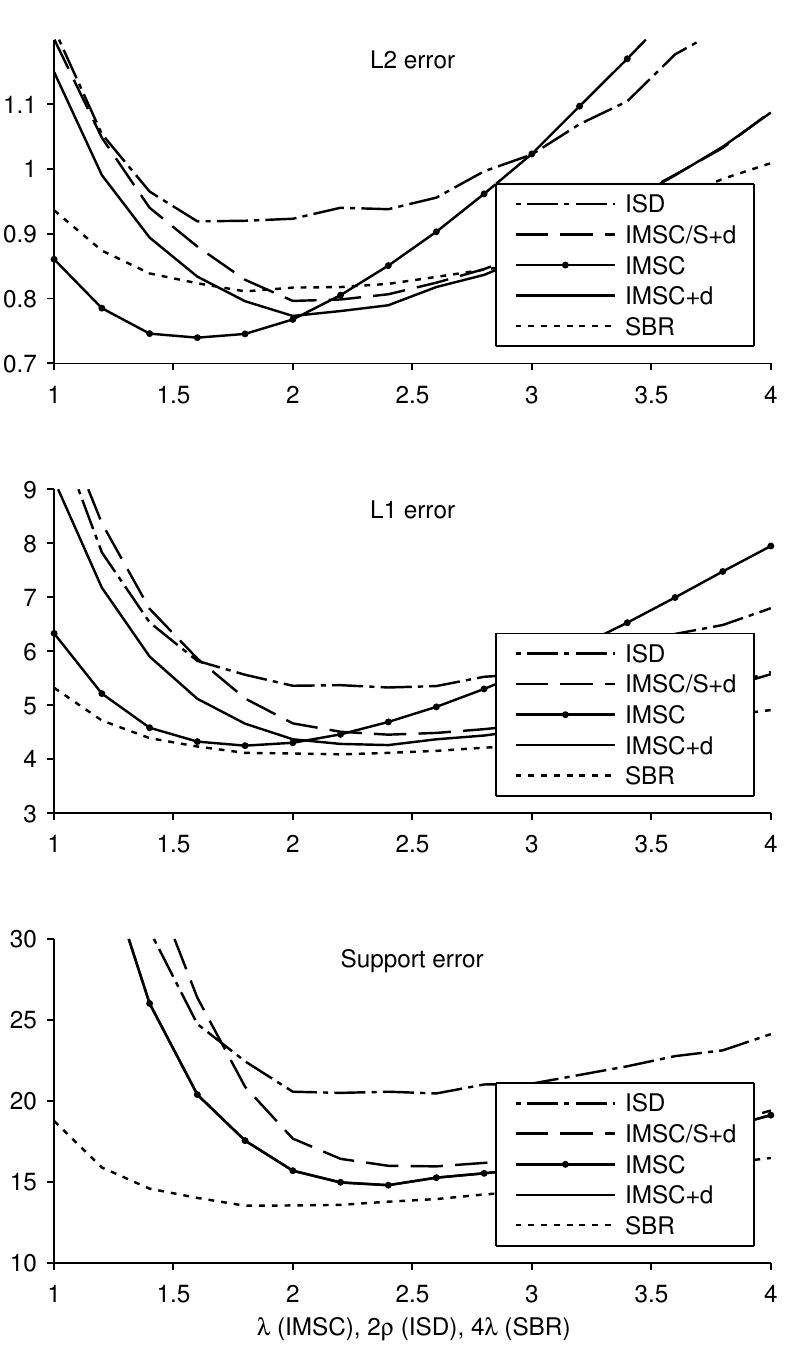}
	\caption{
		Sparse deconvolution.
		Errors as functions of regularization parameters, averaged over 100 realizations.
		(Note that the support error for IMSC and IMSC+d coincide.)
	}
	\label{fig:deconv3}
\end{figure}

Fig.~\ref{fig:deconv3} illustrates the average errors as functions
of the regularization parameter, for ISD, IMSC,
and IMSC + debiasing (denoted IMSC+d in the figure).
For IMSC, the regularization parameter is $\lam$. 
For ISD, the regularization parameter is $\rho = \lam/2$.
Note that for IMSC, the value of $\lam$ minimizing L2E and L1E depends on 
whether or not debiasing is performed.
The value $\lam$ suggested by \eqref{eq:setlamconv} (i.e., $\lam = 2$) 
is reasonably effective with or without debiasing.
The value of $\lam$ minimizing SE is somewhat higher.

The implementation of the $\ell_1$,  IRL2, IRL1, and IMSC algorithms 
for deconvolution each require the solution of \eqref{eq:defF}
with various penalty functions and/or sub-matrices of $\H$.
We have used algorithms, based on majorization of the penalty function,
that exploit banded matrix structures for computational
efficiency \cite{Selesnick_2012_cnx_deconv, Selesnick_2012_cnx_sparse_penalties}.

Finally, we comment on the computation time for IMSC.
The IMSC solution with the atan penalty illustrated in Fig.~\ref{fig:deconv} took 1.7 seconds 
and about 94\% of the time was spent on solving SDPs.
As noted above, three SDPs were solved (of sizes 61, 40, and 38).
The IMSC solution with the log penalty took 2.8 seconds, with again,
about 94\% of the time spent on SDPs.
The longer time was due to more iterations of IMSC (five SDPs instead of three).
The $\ell_1$ norm solution was obtained in only 52 milliseconds (33 times faster than the MSC solution).

\section{Conclusion}

This paper proposes an approach (MSC) to obtain sparse solutions to ill-posed linear inverse problems.
In order to induce sparsity more strongly than the $ \ell_1 $ norm, the 
MSC approach utilizes non-convex penalty functions.
However, the non-convex penalty functions are constrained so that the total cost function is convex. 
This approach was introduced in \cite{Blake_1987},
and extended in \cite{Nikolova_1998_TIP, Nikolova_1999_TIP, Nikolova_2008_SIAM}.
A novelty of the proposed approach is that 
the maximally non-convex (maximally sparsity-inducing) penalty functions
are found by formulating a semidefinite program (SDP).
Iterative MSC (IMSC) consists of applying MSC to the non-zero (active) elements
of the sparse solution produced by the previous iteration.
Each iteration of IMSC involves the solution to a convex optimization problem.

The MSC method is intended as a convex alternative to $\ell_1$ norm minimization,
which is widely used in sparse signal processing where it is often desired that a `sparse' or the `sparsest' solution be found to 
a system of linear equations with noise. 
At the same time, some practitioners are concerned with non-convex optimization issues.
One issue is entrapment of optimization algorithms in local minima.
But another issue related to non-convex optimization is the sensitivity of the solution to perturbations in the data. 
Suppose a non-convex cost function has two minima, one local, one global.
The cost function surface depends on the observed data.
As the observed data vary, the local (non-global) minimum may decrease in value relative to the global minimum.
Hence the global minimizer of the non-convex cost function is a discontinuous function of the data,
i.e., the solution may jump around erratically as a function of observed data.
This phenomena is exhibited, for example, as spurious noise spikes in wavelet hard-thresholding denoising, as illustrated in Fig. ~\ref{fig:dnoise}.
For these reasons, some may favor convex formulations.
The proposed MSC approach simply considers the question: what is the \emph{convex} optimization problem that 
best promotes sparsity (from a parameterized set of penalty functions).

Being based entirely on convex optimization,
it can not be expected that MSC produces solutions as sparse as non-convex optimization methods,
such as $\ell_p$ quasi-norm ($ 0 < p < 1 $) minimization.
However, it provides a principled approach for enhanced sparsity relative to the $\ell_1$ norm.
Moreover, although it is not explored here, 
it may be effective to use MSC in conjunction with other techniques.
As has been recognized in the literature, and as illustrated in the sparse deconvolution example
above, reweighted $\ell_1$ norm minimization can be more effective
than reweighted $\ell_2$ norm minimization (i.e., higher likelihood of 
convergence to a global minimizer).
Likewise, it will be of interest to explore the use of reweighted MSC
or similar methods as a means of more reliable non-convex optimization.
For example,
a non-convex MM-type algorithm may be conceived wherein 
a specified non-convex penalty function 
is majorized  by a non-convex function constrained so as
to ensure convexity of the total cost function at each iteration of MM.

To apply the proposed approach to large scale problems (e.g., image and video reconstruction),
it is beneficial to solve \eqref{eq:sd} by some algorithm 
that does not rely on accessing or manipulating individual rows or columns of $ \H $.

The technique, where a non-convex penalty is chosen so as to lead to a convex problem,
has recently been utilized for group-sparse signal denoising in \cite{Chen_2013_arXiv_MSC}.

\appendix

Suppose $ F $, defined in \eqref{eq:deftf}, is strictly convex
and $\reg(x)$ is differentiable for all $x \in \RR $ except $ x = 0 $.
Then the subdifferential  $ \partial F $ is given by
\begin{equation}
	\label{eq:subdiff}
	\partial F(x) = 
	\begin{cases}
		\{ x-y + \lam \reg'(x) \}, 	&\text{ if } x \neq 0,\\
		[\lam \reg'(0^-), \, \lam \reg'(0^+)]-y,  &\text{ if } x = 0.
	\end{cases}
\end{equation}
Since $ F $ is strictly convex, its minimizer $ x\opt $ satisfies $ 0 \in \partial F(x\opt) $.

If $ y \in [\lam \reg'(0^-), \, \lam \reg'(0^+)]$,
then from \eqref{eq:subdiff} we have $0 \in \partial F(0)$,
and in turn  $ x\opt = 0 $.
Assuming $ \reg $ is symmetric, then $ \reg'(0^-) = -\reg(0^+) $,
and this interval represents the thresholding interval of $\tf$, 
and
the threshold $T$ is given by $ T = \lam \, \reg'(0^+) $.

Suppose now that $y\notin [\lam \reg'(0^-), \, \lam \reg'(0^+)]$. 
This happens if either (i) $y > \lam \reg'(0^+)$, or (ii) $y < \lam \reg'(0^-)$.
In the following, we study case (i).
The results extend to (ii) straightforwardly.

First, note that if $y > \lam \reg'(0^+)$, then $x\opt > 0$ 
and it satisfies
\begin{equation}
	\label{eq:yif}
	y = x\opt + \lam \reg'(x\opt).
\end{equation}
Let us define $ f : \RR_+ \to \RR $ as
\begin{equation}
	\label{eq:deff}
	f(x) = x + \lam \reg'(x).
\end{equation}
Note that, for $x > 0$,  $f(x) = F'(x) + y$.
Since $F(x)$ is strictly convex, $F'(x)$ and $f(x)$ are strictly increasing, hence injective for $x > 0$. 
For $y > \lam \reg'(0^+)$, the threshold function $\tf $ can now be expressed as
\begin{equation}
	\label{eq:tffinv_app}
	\tf(y) = f\inv(y).
\end{equation}
Observe that $f$ is continuous and $f(0^+) = \lam \reg'(0^+) = T$.
In view of \eqref{eq:tffinv_app}, this implies that $\tf(T^+) = 0$.
Thus, $\tf(y)$ is continuous at the threshold.

For a symmetric $\reg$, it can be shown that $ F$ is strictly convex if and only if $ f $ is strictly 
increasing for $x>0$. This in turn can be ensured by requiring $ \reg''(x) > -1/\lam, \  \forall x > 0$.

Let us now find the first and second derivatives of $\tf(y)$ at $ y = T^+ $.
From \eqref{eq:tffinv_app},
$	f(\tf(y)) = y $.
Differentiating with respect to $y$ gives
\begin{equation}
	\label{eq:f1}
	f'(\tf(y)) \, \tf'(y) = 1.
\end{equation}
Differentiating again with respect to $ y $ gives
\begin{equation}
	\label{eq:f2}
	f''(\tf(y)) \, [\tf'(y)]^2 + f'(\tf(y)) \, \tf''(y) = 0.
\end{equation}
Setting $ y = T^+ $ in \eqref{eq:f1} gives
\begin{equation}
	\tf'(T^+) = 1/f'(0^+).
\end{equation}
Setting $ y = T^+ $ in \eqref{eq:f2} gives
\begin{equation}
	f''(0^+) \, [\tf'(T^+)]^2 + f'(0^+) \, \tf''(T^+) = 0	
\end{equation}
or
\begin{equation}
	\tf''(T^+) = -{f''(0^+)}/{[f'(0^+)]^3}.
\end{equation}
Using \eqref{eq:deff}, we have
\begin{equation}
	\label{eq:deffB}
	f'(0^+) = 1 + \lam \reg''(0^+)
	\quad
	\text{and}
	\quad
	f''(0^+) = \lam \reg'''(0^+).
\end{equation}
Equations \eqref{eq:dtf1} and \eqref{eq:dtf2} follow.

\section*{Acknowledgment}

The authors thank an anonymous reviewer for detailed suggestions and corrections that improved the manuscript.

\bibliographystyle{plain}

\end{document}